\pdfoutput=1

\documentclass[11pt]{article}

\usepackage[final]{acl}

\usepackage{times}
\usepackage{latexsym}

\usepackage[T1]{fontenc}

\usepackage[utf8]{inputenc}

\usepackage{microtype}

\usepackage{inconsolata}

\usepackage{graphicx}

\usepackage{booktabs}

\usepackage{float}

\newcommand{\ours}[1]{\textsc{GlobalGroup}}

\DeclareUnicodeCharacter{03BA}{\ensuremath{\kappa}}

\title{Think Globally, Group Locally:\\Evaluating LLMs Using Multi-Lingual Word Grouping Games}

\author{César Guerra-Solano \hspace{10 mm} Zhuochun Li \hspace{10 mm} Xiang Lorraine Li\\
        School of Computing and Information \\ University of Pittsburgh \\ 
  \texttt{\{cguerrasol, zhl163, xianglli\}@pitt.edu} \\}

\begin{document}
\maketitle

\begin{abstract}

Large language models (LLMs) can exhibit biases in reasoning capabilities due to linguistic modality, performing better on tasks in one language versus another, even with similar content.
Most previous works evaluate this through reasoning tasks where reliance on strategies or knowledge can ensure success, such as in commonsense or math tasks. However, abstract reasoning is vital to reasoning for everyday life, where people apply ``out-of-the-box thinking'' to identify and use patterns for solutions, without a reliance on formulaic approaches. Comparatively, little work has evaluated linguistic biases in this task type.
In this paper, we propose a task inspired by the New York Times \textit{Connections}: \ours{}, that evaluates models in an abstract reasoning task across several languages.
We constructed a game benchmark with five linguistic backgrounds -- English, Spanish, Chinese, Hindi, and Arabic -- in both the native language and an English translation for comparison. 
We also proposed game difficulty measurements to evaluate models on games with similar difficulty, enabling a more controlled comparison, which is particularly important in reasoning evaluations.
Through experimentation, we find English modalities largely lead to better performance in this abstract reasoning task, and performance disparities between open- and closed-source models.\footnote{\textbf{Code and data}: \url{https://github.com/cgsol/globalgroup}}

\end{abstract}
\section{Introduction}
\begin{figure}
    \centering
    \includegraphics[width=\columnwidth]{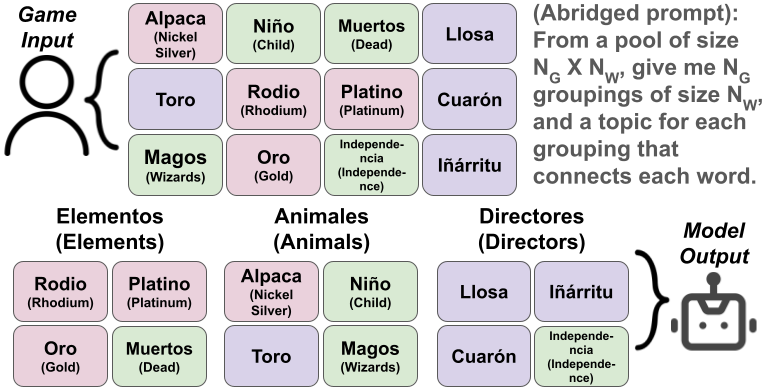}
    \caption{An example Spanish (\textsc{es}) \ours{} game. The model is given 12 words and is instructed to output 3 groups of 4 words, along with the topics it identifies connects each group of words. In the figure, same-colored words form a group. The model made mistakes in its grouping.}
    \label{fig_example_game}
\end{figure}

Abstract reasoning is driven by dynamic thinking, encouraging identifying and building patterns to solve problems, rather than largely relying on previous knowledge or structured solutions.
Large Language Models (LLMs) have demonstrated strong performance in reasoning-centered tasks, where models must reason given a set of information~\cite{deepseekai2025deepseekr1incentivizingreasoningcapability}. This performance can be language-dependent, manifesting as differential task performance depending on the language, even given similar content~\cite{etxaniz-etal-2024-multilingual, ahuja-etal-2023-mega, vayani2025almbench}. However, these investigations have left abstract reasoning tasks relatively unexplored, with a prevalence of English tasks here, presenting a gap for further research~\cite{xu2024llms, huang-etal-2024-lateval}.

An important aspect of abstract reasoning is the ability to understand and utilize knowledge to reason in complex situations, such as reasoning under constraints. This has been explored through work investigating the usage of the New York Times (NYT) \textit{Connections}, a constrained word grouping game, as an abstract reasoning evaluation. The game tasks players to, with a pool of 16 words, find 4 groups of 4 words, each linked under a topic.

While the game-based format affords a unique and customizable test-bed for assessing abstract reasoning capabilities, as the game's format could include different groupings to assess reasoning capabilities relative to different content, the game is only available in English and no work has investigated multilingual derivatives. To bridge these gaps, we introduce a game-based reasoning benchmark to evaluate an LLM's ability when solving abstract reasoning tasks in different languages. 

We propose \ours{}, a novel word grouping game inspired by NYT \textit{Connections}, that evaluates a facet of a model’s abstract reasoning capabilities in different languages. In \ours{}, the LLM is given a pool of words, and with them, must create equal groups of words and provide a topic (oftentimes more abstract than just pertaining to semantic relatedness) that connects each group's words. The model must reason about the given set of words to succeed because of the game's constraints, i.e., limiting word group size. This involves optimizing the word groupings by identifying commonalities among words without interfering with other grouping choices. Due to this game-based format enforcing lateral thinking, such as inductive reasoning via recognizing group patterns, rather than knowledge proficiency or other forms of structured problem-solving, \ours{} assesses abstract reasoning capabilities across languages, addressing a gap in current reasoning evaluation research. Figure~\ref{fig_example_game} presents a Spanish \ours{} example. The dataset contains groupings from 5 languages -- Spanish, Chinese, Hindi, Arabic, and English -- and 511 \textit{Connections} and 600 \textit{Connections}-derived games.

We evaluate four open- and two closed-source models and find:
\begin{itemize}
    \item Across all models, we see a \textbf{bias towards English representations}, represented by increases in performance when translating non-English groupings into English.
    \item By comparing open- and closed-source model results, the \textbf{immense value of a multilingual-focused training paradigm} is shown, as this allows a small open-source model to perform on-par with far larger closed- and open-source LLMs. We additionally note a \textbf{significant contribution of model size} to both overall performance and translation-related performance differences.
    \item Through a game difficulty-based analysis, we identify \textbf{three game features (``difficulty metrics'') correlating with model performance}, hinting at possible game factors impacting model reasoning ability.
\end{itemize}
\section{Related Work}
\label{sec:related work}

Reasoning capabilities of LLMs have become significantly more advanced over time. Recent work has explored evaluating these capabilities, spanning across a variety of reasoning types. For example, mathematical and code reasoning evaluations test skills relating to recognizing and applying formulas or algorithms to achieve a correct answer~\cite{hendrycks2021math, austin2021programsynthesislargelanguage, li2024learning, wang2025performancereviewllmsolving}. 
Commonsense reasoning integrates a variety of reasoning types, with a dependence on prior ``commonsense knowledge''. Many of this commonsense knowledge relies on social customs and norms, which is also the focus of many social reasoning evaluation benchmarks~\cite{talmor-etal-2019-commonsenseqa, sap-etal-2019-social}.

An important reasoning type is abstract reasoning, where people employ skills like abductive or inductive reasoning to identify and build patterns for solutions, rather than relying on existing content, like formulas or specific knowledge. With this, evaluations encourage out-of-the-box thinking and less reliance on pre-set strategies or formulaic approaches~\cite{huang-etal-2024-lateval, xu2024llms}.
While multilingual studies are prevalent in assessing other forms of reasoning~\cite{etxaniz-etal-2024-multilingual, ahuja-etal-2023-mega, vayani2025almbench}, little work has done so for abstract reasoning, with primarily English evaluations here. With the importance of abstract reasoning to everyday life and human intelligence, it is vital that we evaluate the extent of a potential linguistic disparity, informing future developments in creating equitable and diverse models.

\label{connections-full-description}
NYT \textit{Connections}, a game released in June 2023, has recently been used as an abstract reasoning evaluation for LLMs. 
In the game, the player, with a pool of 16 words, must find 4 groups of 4 words, with each group of words connected by a topic, varying in abstractness. These topics relate to domains like alternate word usages, pop culture, morphology, and more. \textit{Connections} games additionally oftentimes have word overlap between groupings (words that could reasonably be put in several prospective groupings in a game), or red herrings (words that could be within a topic/group, but aren't actually, such as ``Sponge'', ``Bob'', ``Square'', and ``Pants''). Its use for this purpose is due to its emphasis on abstract reasoning skill, rather than just knowledge, for success in-game -- the game has users identify and optimize potential groupings, in light of potential red herrings, through a lateral thinking paradigm where pre-set strategies may not lead to success~\cite{merino2024making}. 

Prior work converted the Connection game to a standard English abstract reasoning evaluation task~\cite{samadarshi2024connecting, todd2024missed}.
In these experiments, evaluation was performed using \textit{Connections} games, with no new groupings or games created. These works also do not require the model to produce possible group names, making it difficult to understand rationale behind model predictions.
Additionally, little work has used games (or \textit{Connections} derivatives) as multilingual reasoning evaluations for LLMs. Our work strives to fill this gap, using a multilingual game to assess abstract reasoning.
\section{Dataset}
\label{sec:dataset}
\begin{table*}[]
\centering
\small
\begin{tabular}{@{}ccccccc@{}}
\toprule
\textbf{\begin{tabular}[c]{@{}c@{}}Dataset\\ Background\\ Subsets\end{tabular}} & \textbf{\begin{tabular}[c]{@{}c@{}}Available\\ Languages\end{tabular}} & \textbf{\begin{tabular}[c]{@{}c@{}}Non-Culturally-\\ Related Group Topic\\ (in en)\end{tabular}} & \textbf{\begin{tabular}[c]{@{}c@{}}Culturally-Related (C-R)\\ Group Topic (in en)\end{tabular}} & \textbf{\begin{tabular}[c]{@{}c@{}}C-R\\ Group\\ Amount\end{tabular}} & \textbf{\begin{tabular}[c]{@{}c@{}}Total\\ Group\\ Amount\end{tabular}} & \textbf{\begin{tabular}[c]{@{}c@{}}Game Settings\\ ($N_{words} \times$ \\ $N_{groups}$)\end{tabular}} \\ \midrule
\multicolumn{1}{c|}{en}                                                         & \multicolumn{1}{c|}{en}                                                & Professions                                                                                      & American Holidays                                                                               & 22                                                                    & \multicolumn{1}{c|}{48}                                                 & \begin{tabular}[c]{@{}c@{}}$\{2,3,4\} \times$\\ $\{2,3,4\}$\end{tabular}                                                                          \\ \midrule
\multicolumn{1}{c|}{es}                                                         & \multicolumn{1}{c|}{es, en}                                            & Aquatic Animals                                                                                  & \begin{tabular}[c]{@{}c@{}}Characters From\\ Don Quixote\end{tabular}                           & 27                                                                    & \multicolumn{1}{c|}{48}                                                 & \begin{tabular}[c]{@{}c@{}}$\{2,3,4\} \times$\\ $\{2,3,4\}$\end{tabular}                                                                          \\ \midrule
\multicolumn{1}{c|}{zh}                                                         & \multicolumn{1}{c|}{zh, en}                                            & Tree Composition                                                                                 & The Five Classics                                                                               & 35                                                                    & \multicolumn{1}{c|}{80}                                                 & \begin{tabular}[c]{@{}c@{}}$\{2,3,4\} \times$\\ $\{2,3,4\}$\end{tabular}                                                                          \\ \midrule
\multicolumn{1}{c|}{hi}                                                         & \multicolumn{1}{c|}{hi, en}                                            & Parts of a Train                                                                                 & Fairy Tales                                                                                     & 32                                                                    & \multicolumn{1}{c|}{49}                                                 & \begin{tabular}[c]{@{}c@{}}$\{2,3,4\} \times$\\ $\{2,3,4\}$\end{tabular}                                                                          \\ \midrule
\multicolumn{1}{c|}{ar}                                                         & \multicolumn{1}{c|}{ar, en}                                            & Things That Spin                                                                                 & Traditional Food                                                                                & 19                                                                    & \multicolumn{1}{c|}{40}                                                 & \begin{tabular}[c]{@{}c@{}}$\{2,3,4\} \times$\\ $\{2,3,4\}$\end{tabular}                                                                          \\ \midrule
\multicolumn{1}{c|}{nyt-seq}                                                    & \multicolumn{1}{c|}{en}                                                & Basic Directions                                                                                 & Starts of U.S. Presidents                                                                       & N/A                                                                   & \multicolumn{1}{c|}{2044}                                               & \begin{tabular}[c]{@{}c@{}}$4\times2, 4\times3,$\\ $4 \times 4$\end{tabular}                                                                      \\ \midrule
\multicolumn{1}{c|}{nyt-shuf}                                                   & \multicolumn{1}{c|}{en}                                                & Basic Directions                                                                                 & Starts of U.S. Presidents                                                                       & N/A                                                                   & \multicolumn{1}{c|}{2044}                                               & \begin{tabular}[c]{@{}c@{}}$\{2,3,4\} \times$\\ $\{2,3,4\}$\end{tabular}                                                                          \\ \bottomrule
\end{tabular}

\caption{Dataset Overview: Our dataset includes games from five languages, with 11 subsets, including the English translations of \textsc{es}, \textsc{zh}, \textsc{hi}, and \textsc{ar}. The \textsc{nyt-seq} and \textsc{nyt-shuf} dataset is sourced from the New York Times Connections, separated due to different game creation procedures. We provide examples of both non-culturally-related and culturally-related groups to show the dataset diversity. Except for \textsc{nyt-seq}, which has three game settings, all other subsets have nine game settings. Additional dataset examples can be found in Appendix~\ref{app:example-games}.}
\label{tab:total-dataset-overview}
\end{table*}

\ours{} has a pool of $x$ words, with the game goal being to form $m$ groups of $n$ words where $m\times n=x$. Words within the same group share a common topic or category, distinguishing them from words in other groups. Finding a solution to the game requires a reasoning process, which we aim to evaluate in different languages. Models are prompted with the word pool and instructed to return groupings as previously specified and topics connecting each grouping's words.

We create \ours{} games by sampling $m$ groupings from a language's word grouping dataset, and sampling $n$ words from each grouping to create a game with a pool of $x$ words.

\subsection{Word Groupings}
We create 6 word groupings datasets -- English (\textsc{en}), Spanish (\textsc{es}), Chinese (\textsc{zh}), Hindi (\textsc{hi}), Arabic (\textsc{ar}), and compiled groupings from 511 \textit{Connections} games (\textsc{nyt}). To create the grouping dataset for a language, native speaker annotators are asked to create groups of 4 words connected by a topic. Due to the prevalence in \textit{Connections} of culturally-related groupings (with topics pertaining to aspects of culture in English-speaking regions, such as slang, cultural dishes, or pop culture) and non-culturally-related groupings (with topics less pertaining to aspects of culture, such as animals), annotators are asked to generate groups with culturally-related and non-culturally-related topics relative to their specific language to mirror this quality, labeling groups as such. The authors then agree on these labels. Annotator background information can be found in Appendix~\ref{app:annotator-information}.

Following a quality check of annotator groupings\footnote{The quality check included ensuring that the groupings consist of words connected by a topic, and that topics in a groupings dataset do not repeat.}, this resulted in 48 \textsc{en}, 48 \textsc{es}, 80 \textsc{zh} groupings, 49 \textsc{hi}, and 40 \textsc{ar} groupings. We additionally translate non-English groupings into English, creating four additional groupings datasets: \textsc{es-en} (\textsc{es} in English), \textsc{zh-en} (\textsc{zh} in English), \textsc{hi-en} (\textsc{hi} in English), and \textsc{ar-en} (\textsc{ar} in English). These translations were performed by the annotators, with verification assistance from online translation tools Google Translate\footnote{\url{https://translate.google.com/}} and DeepL\footnote{\url{https://www.deepl.com/en/translator}}. Further details on the translation protocol are in Appendix~\ref{app:annotator-information}.

\subsection{The New York Times Connection Game}
Additionally, a dataset of groupings from \textit{Connections} was collected from an online guide\footnote{\url{https://tryhardguides.com/nyt-connections-answers/}}, featuring all groupings across 511 \textit{Connections} games (2,044 groupings total). These groupings were used as sources for \ours{} games via random sampling.
These were included in experimentation to compare performance between author-created \ours{} groups and games (especially in \textsc{en}) to existing \textit{Connections} groups and games.

\subsection{\ours{}}
The default \ours{} game consists of 16 words forming four groups of four. To evaluate model performance across various game settings, we also created games with 2, 3, and 4 groups, each containing 2, 3, or 4 words per group. These variations were generated through random sampling without replacement, resulting in 9 experimental settings for \textsc{en}, \textsc{es}, \textsc{es-en}, \textsc{zh}, \textsc{zh-en}, \textsc{hi}, \textsc{hi-en}, \textsc{ar}, and \textsc{ar-en}. For each setting, 600 games were created via sampling and then evenly split into development and test sets with 300 games for each set. The authors further ensured that no repeating words or group topics appeared within a game, guaranteeing that each game had a single correct solution. This is further discussed in Appendix~\ref{app:multiple-solutions}.

Additionally, to compare performance between author-created \ours{} groups and games and \textit{Connections} groups and games, \textit{Connections} groupings from 511 games were also used. These games all have 4 groupings per game. To create diverse experiments, words per grouping in \textit{Connections} games were randomly sampled to create \textsc{nyt-seq}, resulting in 3 experimental settings (2-, 3-, and 4-word 4-group games) with 211 and 300 games in the development and test sets respectively.

Intuitively, words in the \textit{Connections} game have the potential to count for multiple groupings (\textit{overlap} between groupings), which can make the task more difficult as players need to consider multiple group possibilities at once~\cite{merino2024making}. This notion of word overlap is explicitly used to create challenging \textit{Connections} games~\cite{liuconnections2023}. Thus, we hypothesize that purposeful word overlap between groupings from \textit{Connections} games increases difficulty.
To evaluate this, we also randomly sampled groups and words from the original games to create \textsc{nyt-shuf}, resulting in 9 experimental settings (2-, 3-, and 4-word 2-, 3-, and 4-group games) with 600 games evenly split into development and test sets, providing an English subset similar to other language subsets in the benchmark.
\section{Experiments}
\label{Experiments}
For experimentation, we evaluate closed- and open-source models on 4-group, 4-word games for each subset in \ours{}. Section~\ref{sec:evaluation} explains the evaluation metrics for all games. Section~\ref{sec:difficulty analysis} describes how we measure game difficulty and Section~\ref{sec:baseline} provides details on the baseline models.

\subsection{Evaluation}
\label{sec:evaluation}
For evaluation, a model is provided with the \ours{} game premise, the shuffled word pool for a given game, and instructions regarding the output format. The model is instructed to output several groups of words, along with an associated topic for each group. The exact prompts used can be seen in Appendix~\ref{app:prompts}. We evaluate the model's performance based on both the predicted groups and topics.

\paragraph{Matching Attempts to Ground Truth Groups}
As the models are not required to output their responses in a particular order associated with the true answer, matching the model's predicted groups to the ground truth groups can be challenging. We use an intuitive strategy: true groupings are assigned to the predicted groupings with the greatest set intersection, referred to as ``attempt groups.'' If the number of predicted groups is fewer than the ground-truth groups, the unmatched true groupings are mapped to \textsc{NULL} and scored as complete misses. If the model predicts too many groups, the remaining predicted groups after group mapping are ignored from evaluation. We do so, rather than penalizing for additional group responses, to accommodate the potential for extra groupings outside of lapses in reasoning, such as answer explanations. This is further discussed in Appendix~\ref{app:additional-groups-in-responses}.

\paragraph{Grouping Evaluation} 
We use \textbf{group-level F1} to compare each matched pair of predicted and true groups. For each word in a ground-truth group, a positive point is defined as the word appearing in the attempted group; otherwise, it is negative. The F1 score for each game is then averaged across groups for a game. Unlike metrics used in works that evaluate with \textit{Connections}, such as the unweighted clustering score~\cite{samadarshi2024connecting}, which considers a predicted group correct only if all words are correct (further referred to as "CTD"), group-level F1 allows a granular analysis of model performance, with model predictions evaluated on a scale based on their degree of correctness. We test and support this hypothesis in Appendix~\ref{app:group-evaluation-metrics}.

\paragraph{Topic Evaluation} 
We also evaluate the predicted group topics using a \textbf{"Topic Achieved" (TA)} score, a boolean indicating if a topic was successfully predicted. The TA score is calculated by comparing predicted and ground-truth topics using FastText embeddings~\cite{fasttext-bojanowski-etal-2017-enriching}. We compute the cosine similarity between the predicted topic and its matched true topic, as well as the other true topics in the same game. If the predicted topic and its matched true topic have a similarity $\ge$ 0.3, and is more similar than the predicted topic and the other true topics, a TA score of 1 is given. We report the success rate of topic prediction.

We experimented with both BERTScore~\cite{bert-score} and FastText embeddings with cosine similarity as the basis for the TA score, testing a range of similarity thresholds from 0.1 to 0.7 in 0.1 increments. To determine the best setting, two human annotators annotated the topic achievement of 200 Chinese grouping attempts. We calculated Randolph's Kappa between the annotators, treating automatic predictions (either BERTScore or FastText similarity) as one of the annotators \cite{randolphskappa}. From this, we found that a FastText embeddings-based approach with a similarity threshold of 0.3 performed best, attaining a Randolph's Kappa score of 0.55. 
Further details are in Appendix~\ref{app:determining-ta}.

\begin{table*}[]
\centering
\small
\begin{tabular}{@{}cccccllllccc@{}}
\toprule
\textbf{Models}                        & \multicolumn{11}{c}{\textbf{F1 Score}}                                                                                                                                                                                                              \\ \midrule
\multicolumn{1}{c|}{\textit{Datasets}} & es             & es-en          & zh             & zh-en          & \multicolumn{1}{c}{hi} & \multicolumn{1}{c}{hi-en} & \multicolumn{1}{c}{ar} & \multicolumn{1}{c}{ar-en} & \multicolumn{1}{c|}{en}             & nyt shuf.      & nyt seq.       \\ \midrule
\multicolumn{1}{c|}{GPT-3.5-Turbo}     & 0.828          & 0.855          & 0.927          & 0.836          & 0.886                  & 0.948                     & 0.756                  & 0.916                     & \multicolumn{1}{c|}{0.874}          & 0.594          & 0.531          \\
\multicolumn{1}{c|}{GPT-4}             & \textbf{0.892} & \textbf{0.920} & \textbf{0.971} & \textbf{0.935} & \textbf{0.963}         & \textbf{0.964}            & \textbf{0.877}         & \textbf{0.957}            & \multicolumn{1}{c|}{\textbf{0.943}} & \textbf{0.813} & \textbf{0.707} \\ \midrule
\multicolumn{1}{c|}{Llama3-8B}         & 0.676          & 0.743          & 0.631          & 0.849          & 0.724                  & 0.878                     & \textbf{0.829}         & 0.815                     & \multicolumn{1}{c|}{0.776}          & 0.606          & 0.610          \\
\multicolumn{1}{c|}{Llama3.1-70B}      & \textbf{0.830} & \textbf{0.860} & \textbf{0.954} & \textbf{0.944} & \textbf{0.942}         & \textbf{0.957}            & 0.823                  & \textbf{0.916}            & \multicolumn{1}{c|}{\textbf{0.886}} & \textbf{0.708} & 0.642          \\
\multicolumn{1}{c|}{Mistral-7B}        & 0.565          & 0.677          & 0.771          & 0.825          & 0.497                  & 0.927                     & 0.612                  & 0.789                     & \multicolumn{1}{c|}{0.790}          & 0.584          & 0.577          \\
\multicolumn{1}{c|}{Aya-8B}            & 0.745          & 0.794          & 0.884          & 0.843          & 0.873                  & 0.940                     & 0.734                  & 0.814                     & \multicolumn{1}{c|}{0.822}          & 0.656          & \textbf{0.672} \\ \midrule
\multicolumn{1}{c|}{}                  & \multicolumn{11}{c}{\textbf{\% FastText Topic Achieved (threshold = 0.3)}}                                                                                                                                                                          \\ \midrule
\multicolumn{1}{c|}{GPT-3.5-Turbo}     & 0.597          & 0.693          & 0.661          & 0.706          & 0.626                  & 0.755                     & 0.495                  & 0.648                     & \multicolumn{1}{c|}{0.700}          & 0.312          & 0.288          \\
\multicolumn{1}{c|}{GPT-4}             & \textbf{0.663} & \textbf{0.725} & \textbf{0.721} & \textbf{0.712} & \textbf{0.775}         & \textbf{0.790}            & \textbf{0.622}         & \textbf{0.792}            & \multicolumn{1}{c|}{\textbf{0.758}} & \textbf{0.463} & \textbf{0.407} \\ \midrule
\multicolumn{1}{c|}{Llama3-8B}         & 0.418          & 0.543          & 0.434          & 0.661          & 0.346                  & 0.676                     & \textbf{0.548}         & 0.563                     & \multicolumn{1}{c|}{0.554}          & 0.293          & 0.280          \\
\multicolumn{1}{c|}{Llama3.1-70B}      & \textbf{0.622} &\textbf{ 0.697} & \textbf{0.691} & \textbf{0.719} & \textbf{0.673}         & \textbf{0.785}            & 0.541                  & \textbf{0.658}            & \multicolumn{1}{c|}{\textbf{0.705}} & \textbf{0.387} & \textbf{0.338} \\
\multicolumn{1}{c|}{Mistral-7B}        & 0.257          & 0.550          & 0.448          & 0.650          & 0.296                  & 0.744                     & 0.429                  & 0.562                     & \multicolumn{1}{c|}{0.572}          & 0.288          & 0.296          \\
\multicolumn{1}{c|}{Aya-8B}            & 0.445          & 0.507          & 0.596          & 0.578          & 0.572                  & 0.685                     & 0.437                  & 0.481                     & \multicolumn{1}{c|}{0.522}          & 0.244          & 0.228          \\ \bottomrule
\end{tabular}
\caption{The averaged results by model across each 4-group, 4-word game used during experimentation. Results are divided by each dataset and between closed- and open-source models. A bar plot is in Appendix~\ref{fig_4g4w_raw_results_bar_plot}.}
\label{tab:4g4w-raw-results}
\end{table*}

\subsection{Game Difficulty Measurement}
\label{sec:difficulty analysis}
Linguistic modality is not the only factor influencing model performance. Due to the nature of any game, varying difficulty levels (like differing game factors) can impact performance. Therefore, we experimented with different difficulty metrics for \ours{} games. By classifying game difficulties, we can compare model performance across languages while controlling for these difficulty metrics, allowing finer-grained reasoning analysis which is often coveted when evaluating reasoning in LLMs.
Possible sources of game difficulty include handling larger amounts of information, such as having more words to group, or unusual groupings, like dissimilar words in the same group. Other challenges arise from the game's structure -- perceived word overlap between groups (where a word could belong to several topics) has been identified as a significant contributor to difficulty in \textit{Connections}~\cite{merino2024making, liuconnections2023}. We explore all of these factors as explained below. We experimented with the difficulty measurements in Section~\ref{sec:difficulty_results}.

\paragraph{Group Size}
We hypothesize that game size is a strong contributor to game difficulty. We change game size by varying the number of word groupings in a game, and words per grouping in a game.

\paragraph{Clusters by Word Embeddings}
Another source of difficulty could arise when dissimilar words belong to the same group, meaning group word connectedness is less tied to semantic meaning. To verify this, we represent word similarity within a topic by mapping true word groupings to groups formed by clustering words using semantic embeddings (creating groups connected by semantic similarity) and evaluate inter-group similarity using the Adjusted Rand Index (ARI)~\citep{hubert-adjusted-rand-index}. Words within each grouping are represented by their embedding, calculated by tokenizing each word or phrase and averaging the embeddings across the tokens.
Groupings with poor ARI scores relative to their matched similarity-based grouping indicate low semantic similarity between words potentially making the game more difficult.

\paragraph{Word Overlap between Groups}
The final difficulty measurement we explored extends beyond individual groupings in the game. We hypothesize that "word overlap" between groupings, represented by words that could fall under several prospective group topics, is a reasonable predictor of difficulty. This has been confirmed as a significant contributor to difficulty in \textit{Connections}, where word overlap is considered a desirable feature for creating a challenging game~\cite{liuconnections2023} and is used in modeling custom game creation~\cite{merino2024making}. To measure word overlap in the game, we ask GPT-4o-mini to propose grouping candidates without the game group size constraints, and then compute the average group intersection (via set intersection) across these proposed groupings for a game, giving each game a word overlap score. This is further discussed in Section~~\ref{pgraph:word_overlap}.

\subsection{Baselines}
\label{sec:baseline}

We experimented with zero-shot prompting for multiple models. The prompts include the rules of the game and the pool of words for the given game. Additionally, within the prompt, the output format is specified with an example to support the parsing of LLM answers. We experimented with 11 subsets of the datasets for models using each game in each respective dev dataset and reported the results on the test subset. Model responses were parsed using regular expressions and then organized, mapping guessed groupings to guessed topics. During this process, if any game caused issues with answer parsing due to bad response formatting, the model response was reformatted via prompting of GPT-4o-mini. Please find the exact prompts used across experimentation in Appendix~\ref{app:prompts}.

\paragraph{Models} 
Six models, including both closed-source and open-source models, are experimented with using the benchmark. For closed-source LLMs, we utilized OpenAI's GPT-3.5-Turbo \cite{brown2020language} and GPT-4~\cite{openai2024gpt4}.
For open-source LLMs, we utilized Meta's Llama-3-8B-Instruct and Llama-3.1-70B-Instruct~\cite{dubey2024llama} and Mistral AI's Mistral-7B-Instruct-v0.2~\cite{jiang2023mistral}. These three models are three of the highest-performing open-source LLMs at the time of writing, with good performance in various NLP tasks such as reasoning, mathematics, and code generation~\cite{xu2024towards}. This selection of models additionally allows us to observe potential scaling effects, with two different sizes of the same model.
An additional open-source multilingual LLM, Cohere Labs' Aya-23-8B~\cite{aryabumi2024aya}, was also used in the experimentation due to its multilingual focus\footnote{We have tested other open-source and multilingual LLMs: Llama2-7B-chat~\cite{touvron2023llama}, Apollo-7B~\cite{wang2024apollo}, PolyLM-13b~\cite{wei2023polylm}, and Bloomz-7b1~\cite{muennighoff2022crosslingual}. However, these LLMs can't understand our tasks and give meaningful output. Therefore, their results are not included.}. 
More details of the model parameters are shown in Appendix~\ref{app:parameter}. These models were chosen not only due to their high performance across a variety of benchmarks, but also to offer a comparison between privately-owned LLMs of larger size, versus smaller, open-source models.

\section{Results}

We report the model performance across models and subsets of 4-group, 4-word ($4\times4$) games in the benchmark in Table~\ref{tab:4g4w-raw-results}. Additional analyses, omitted due to spatial constraints, can be seen in Appendix~\ref{app:section-additional-analyses}.

\paragraph{Performance with Language of Origin}

Considering both the F1 and TA scores, we can see in Table \ref{tab:4g4w-raw-results} that \textsc{es}, \textsc{hi}, and \textsc{ar} groupings performed significantly better when translated to English across all models. \textsc{zh} groupings had a differing relationship, with better performance in Chinese rather than English translations in closed-source models, but English better than Chinese in open-source models, except for Aya-8b. With the multilingual focus of Aya-8b's training~\cite{aryabumi2024aya}, we assume this differing pattern in performance is due to larger amounts of Chinese text training data. This highlights a strong bias towards English language representations in these models, reflecting limited multilingual reasoning capabilities even for the same knowledge, and stresses the need for diverse language choices when developing training paradigms and benchmarks. An additional figure depicting this trend can be seen in Figure~\ref{fig:f1_difference_due_to_translation}.

\paragraph{Performance with Culture of Origin}
\begin{table}[]
\centering
\tiny
\begin{tabular}{@{}ccccccc@{}}
\toprule
\multicolumn{5}{c}{\textbf{\begin{tabular}[c]{@{}c@{}}Mean F1 Score Difference (Non-Culturally-Related - Culturally-Related)\\ Across All Dataset Subsets\end{tabular}}} \\ \midrule
\multicolumn{1}{c|}{\textbf{Models}}                & GPT-3.5-Turbo                & GPT-4                      &                              &                         \\ \midrule
\multicolumn{1}{c|}{Mean}                           & \textbf{0.050}               & 0.032                      &                              & \textbf{}               \\ \midrule
\multicolumn{1}{c|}{\textbf{Models}}                & Llama3-8B                    & Llama3.1-70B               & Mistral-7B                   & Aya-8B                  \\ \midrule
\multicolumn{1}{c|}{Mean}                           & 0.044                        & 0.045                      & \textbf{0.086}               & 0.065                   \\ \bottomrule
\end{tabular}
\caption{A table depicting the average difference in performance in groups labeled as non-culturally-related versus culturally-related in $4\times4$ games averaged across all dataset subsets (excluding \textsc{nyt-seq} and \textsc{nyt-shuf}) for open- and closed-source models.}
\label{tab:mean-difference-cultural-groups}
\end{table}

Utilizing the culture-relatedness judgment for each grouping from its respective annotator, we investigate the average F1 performance on culturally-related and non-culturally-related groupings, across all datasets. A seen in Table~\ref{tab:mean-difference-cultural-groups}, we find that, when compared to culturally-related groupings, non-culturally-related groupings lead to average F1 performance improvements across all models.
This substantiates a potential cultural basis to model performance and reasoning capabilities as well. The full results, including the performance difference by dataset, can be seen in Table~\ref{tab:mean-difference-cultural-groups-by-dataset}.

\paragraph{Crowd-created Games vs \textit{Connections} Games}

In Table~\ref{tab:4g4w-raw-results}, the first 9 columns represent crowd-created \ours{} games, while the last 2 columns show games derived from the NYT \textit{Connections} games. We can see that performance on the NYT games is significantly lower across all models, particularly for closed-source models when evaluated using the F1 score, and all models when evaluated with the TA score, suggesting that the NYT \textit{Connections} game may be more challenging, especially when compared to the in-house \textsc{en} game. By stratifying \textit{Connections} results by the \textit{Connections}-set group difficulty, we find our \textsc{en} games are similar in performance to ``Yellow'' \textit{Connections} groupings, the easiest connection grouping type. Further analysis is in Appendix~\ref{app:connection-difficulty}.

When comparing the two variations of the Connections game, we also observed a performance drop for \textsc{nyt-seq}, potentially indicating intentionally difficult game design. This is possibly due to intentional group overlap, as that is the only difference between \textsc{nyt-seq} and \textsc{nyt-shuf}, due to the random sampling approach used for \textsc{nyt-shuf}.

\paragraph{Closed-Source vs Open-Source Model Results}
Throughout the experiments, one can note a clear grouping of models with respect to performance -- the smaller, closed-source LLMs Llama3-8B and Mistral-7B perform similarly badly compared to all other models, even considering Aya-8B, a open-source model of similar size. This demonstrates the value of a multilingual-focused training paradigm, as Aya-8B is able to attain comparable results as the far larger LLMs, despite having a substantially smaller size. This can be seen across all scored metrics, and across all dataset backgrounds, pointing to a clear difference in these models. We hope this informs future model development efforts that seek to efficiently adapt to a variety of linguistic and cultural settings.

Additionally, we can note that model size has a large impact on performance. When comparing results from Llama3-8B to Llama3.1-70B, we can note a general increase in performance, with many subsets now reaching $>90\%$ F1 score. We can additionally see how model size impacts performance differences relative to translation, as there are considerably smaller performance deficits between English/non-English subsets, pointing to model size directly impacting the extent of linguistic modality-related performance biases.

While the performance gap between open- and closed-source models (aside from Llama3.1-70B) is large in English-based games, this performance gap is far smaller for non-English-based games, such as Arabic games, contrary to initial expectations. This can inform future work relating to LLM training, and offers a perspective on what tasks can smaller, open LLMs, perform on par with larger, oftentimes cost-intensive, closed LLMs. In this case, we see that these smaller, open LLMs can perform similarly to the larger, closed LLMs on Arabic tasks.

\begin{figure}
    \centering
    \includegraphics[width=\columnwidth]{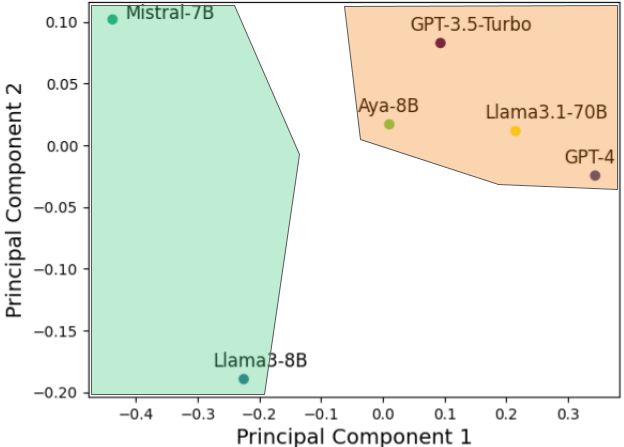}
    \caption{PCA plot of model representations based on their performance. Each point corresponds to a model, projected onto the first two principal components.}
    \label{fig:model_pca}
\end{figure}

We additionally perform a principal component analysis, where each model is represented by a vector of their mean $4\times4$ game results by dataset subset, plotting the first two principal components for each model. As seen in Figure \ref{fig:model_pca}, the closed-source models are clustered together while the open-source models form another cluster, additionally seen in K-means clustering with $k = 2$. Investigating the contributions of each dataset to the first two principal components, we find that \textsc{hi}, \textsc{zh}, and \textsc{ar} have the largest weights, with all of these being the non-Latin script dataset subsets.
\section{Difficulty-Aware Benchmark Results}
\label{sec:difficulty_results}
Model performance on each subset of the game depends on several factors, such as linguistic/cultural differences and game difficulty, which involves both the complexity within the groups and the difficulty across groups within the game. In order to provide granularity in model evaluation, through using games of a specific ``difficulty level'', and control for potential confounding factors to better identify language-related differences in performance, we propose several measures to assess game difficulty.
We
evaluate the effectiveness of these measures in this section. Additional analyses, such as correlation calculations and categorized difficulty results, can be seen in Appendix~\ref{app:additional_difficulty_metric_analysis}.

A good difficulty measure should correlate well with model performance, i.e., the harder the game, the worse the model's performance when game difficulty (agnostic of language or cultural setting) is the only difference between games. Following this intuition, we calculate model performance results for different difficulty metric slices of the benchmark, such as varying group counts or group sizes in the game. For
ARI score and word overlap, we bin them into 3-4
categories and calculate the average performance by bin. The results for GPT-4 and GPT-3.5-Turbo on all game sizes are in Table~\ref{tab:background-averaged-bucket-results}.

We additionally use these difficulty metrics for controlled comparisons across models and languages, although due to spatial constraints, that has been moved to Appendix~\ref{app:additional_difficulty_metric_analysis}.

\begin{table}[]
\centering
\tiny
\begin{tabular}{@{}ccccc@{}}
\toprule
\textbf{Models}                                                                                          & \multicolumn{4}{c}{\textbf{Average F1 Score}}                            \\ \midrule
\multicolumn{1}{c|}{\textit{Group Size}}                                                                 & 2               & 3              & 4              & \multicolumn{1}{l}{} \\ \midrule
\multicolumn{1}{c|}{GPT-3.5-Turbo}                                                                       & 0.881           & \textbf{0.890} & 0.884          &                      \\
\multicolumn{1}{c|}{GPT-4}                                                                               & 0.925           & 0.940          & \textbf{0.949} &                      \\ \midrule
\multicolumn{1}{c|}{\textit{Group Count}}                                                                & 2               & 3              & 4              &                      \\ \midrule
\multicolumn{1}{c|}{GPT-3.5-Turbo}                                                                       & \textbf{0.930}  & 0.883          & 0.842          &                      \\
\multicolumn{1}{c|}{GPT-4}                                                                               & \textbf{0.962}  & 0.940          & 0.912          &                      \\ \midrule
\multicolumn{1}{c|}{\textit{\begin{tabular}[c]{@{}c@{}}Adjusted Rand\\ Index Range\end{tabular}}}        & {[}-0.5, 0.0{]} & (0.0, 0.5{]}   & (0.5, 1.0{]}   &                      \\ \midrule
\multicolumn{1}{c|}{GPT-3.5-Turbo}                                                                       & 0.924           & 0.927          & \textbf{0.951} & \textbf{}            \\
\multicolumn{1}{c|}{GPT-4}                                                                               & 0.969           & 0.974          & \textbf{0.990} & \textbf{}            \\ \midrule
\multicolumn{1}{c|}{\textit{\begin{tabular}[c]{@{}c@{}}Candidate Proposal\\ Overlap Range\end{tabular}}} & {[}0.0, 0.75{]} & (0.75, 1.5{]}  & (1.5, 2.25{]}  & (2.25, 3.0{]}        \\ \midrule
\multicolumn{1}{c|}{GPT-3.5-Turbo}                                                                       & \textbf{0.932}  & 0.873          & 0.813          & 0.831                \\
\multicolumn{1}{c|}{GPT-4}                                                                               & \textbf{0.977}  & 0.923          & 0.863          & 0.859                \\ \bottomrule
\end{tabular}
\caption{The averaged F1 score for closed-source models across each 4-word game (4-group for group size analysis) whose group size, group count, ARI, and Word Overlap score fell within the specified ranges. Results are divided by 3-4 ranges, and averaged across all dataset backgrounds, excluding \textsc{nyt-seq}.}
\label{tab:background-averaged-bucket-results}
\end{table}

\paragraph{Group Size \& Group Count} 
We report results from 4-group games for analyzing the impact of group size, and 4-word groups for analyzing the impact of group count.
We hypothesize that the number of words in each group is associated with game difficulty. From a player's perspective, increasing the group size or context may offer additional opportunities to identify group topics, 
making the game easier. However, as shown in Table \ref{tab:background-averaged-bucket-results}, the observed patterns are inconsistent. While performance clearly increases in games as group size increases for GPT-4, there is an inconsistent pattern for GPT-3.5-Turbo, with performance unusually peaking with a group size of 3, indicating that grouping size may not be a good predictor of game difficulty.
In contrast, varying the number of groups reveals clear patterns in model performance. This is reflected with the average performance relative to the group count of games considerably declining with an increase in group count for both models. Thus, we adopt game group count as one of the main difficulty measurements.

\paragraph{Adjusted Rand Index}

We represent words in each ground-truth group using FastText embeddings and apply K-means clustering, setting $k$ to the number of groups in a game. We use 2, 3, and 4-group games with a group size of 4, giving us 30 subsets across the 10 benchmark subsets. For each subset, we evaluate the clustering results with the Adjusted Rand Index (ARI), binning the results into three categories. We hypothesize that a higher ARI indicates greater semantic similarity within the groups, making the game easier. To test this hypothesis, we compare model performance in games within 3 different bins of ARI values. As shown in Table \ref{tab:background-averaged-bucket-results}, both GPT-4 and GPT-3.5-Turbo exhibit very strong trends, with model performance consistently improving with higher ARI scores. We conclude that increased group inter-word similarity is associated with better performance.

\paragraph{Word Overlap Assessment}
\label{pgraph:word_overlap}
To calculate word overlap, we conducted candidate proposal experiments. Here, we prompted GPT-4o-mini with the word pool of a game and the final group topics, asking it to assign words to topics, with replacement. By allowing words to be assigned to multiple topics, we capture the word overlap between groups in-game. We calculate the overlap by parsing each list of words per topic and computing the set intersection between the lists. The average number of overlapped words in a given game is then calculated by averaging this value across all topics. To validate GPT-4o-mini's performance in this task, a subset of candidate proposal assignments for \textsc{en} were cross-checked by the authors. We observe only valid responses and no illogically overlapped words, therefore we proceed with this experimentation. We then follow the same experimental setup to calculate the average F1 score in games within 4 bins of overlap values, as done in the clustering experiment. As shown in Table \ref{tab:background-averaged-bucket-results}, word overlap proves to be associated with performance, with performance trending as expected, decreasing as word overlap increases. Consequently, we conclude that word overlap is also associated with game difficulty. 
\section{Conclusion}
We have presented a new multilingual dataset, \ours{}, to assess LLM abstract reasoning across languages through a novel, game-based perspective. By having game mechanics which require high performance in abstract reasoning/lateral thinking (encouraging inductive reasoning and pattern recognition), we have created an evaluation method that can assess this trait across multiple languages. As evidenced through experimentation, \ours{} can be used to identify language-related bias, seen through performance increases with English translations, and note consistent performance differences between the closed and open-source models that were tested, observing the value of a multilingual-focused training paradigm for smaller models. Additionally, we provide three difficulty metrics for predicting game difficulty for a model, reflecting previous judgments of aspects contributing to game difficulty for humans. With this benchmark, we provide an avenue for evaluating LLM reasoning abilities across languages, providing a valuable means for assessing linguistic bias within LLMs -- a key component of creating diverse and equitable language systems.

\section{Limitations and Ethical Considerations}

In this paper, we only cover five languages, while recognizing that there are thousands of languages in the world that could be studied, along with a multitude of cultural backgrounds that are not represented. We encourage community members to add to our dataset using our explicit annotator request and dataset construction procedures. 

For the language we covered in the paper, we recognize the vast and diverse nature of language and culture. Since our annotators do not reflect the complete demographics of speakers of a given language, some form of bias may have been introduced. We also acknowledge that culture is a complex term that involves different levels of communities and that our approximation of cultural-relatedness may not fully encompass the vastness of what culture represents. Another limitation is that while the authors filtered for unethical or harmful content in the word groups, while very unlikely, there is a possibility that we might have overlooked some.

\section*{Acknowledgments}

We would like to thank the following entities for their gracious and continual support of our work through resources, feedback, and discussion:
\begin{itemize}
    \item This research was supported in part by the University of Pittsburgh Center for Research Computing and Data, RRID\:SCR\_022735, through the resources provided. Specifically, this work used the HTC cluster, which is supported by NIH award number S10OD028483.
    \item The PittNLP community, for consistent help in data annotation and thoughtful feedback throughout the development of this project.
    \item All annotators, including members of the PittNLP community, the Durrant Lab, the Dietrich School of Arts and Sciences, and community members outside of the University of Pittsburgh, for incredible and diligent work with annotation and translation.
    \item The Stamps Foundation, for providing scholarship support and additional funding throughout the project for the first author.
    \item We also thank Allyson Ettinger and Vered Shwartz for the fruitful discussion and feedback for the project. 
\end{itemize}

\bibliography{custom}

\appendix
\section{Prompts}
\label{app:prompts}
\subsection{Model Response Prompting}
From these prompts, "\#*\#" is replaced with the total number of words in the pool, \$*\$ is replaced with the total number of groups in the game, and {} is replaced with the provided pool of words for a given game. A new prompt is generated for each new game, with each game's metrics being used to fill these spaces in the prompts.

English prompt for 2-group games: \\
I am going to give you a pool of \#*\# words. These \#*\# words can be separated into \$*\$ equal groups of 4 words linked under some category. I want you to tell me the four groups and what category you think connects them. Here is an example: Given the pool ['Mile', 'League', 'Jazz', 'Heat', 'Yard', 'Cabaret', 'Carousel', 'Nets', 'Gobble', 'Scarf', 'Foot', 'Bucks', 'Chow', 'Wolf', 'Cats', 'Chicago'], you would output: <NBA TEAMS>: ['Bucks', 'Heat', 'Jazz', 'Nets'], <UNITS OF LENGTH>: ['Foot', 'League', 'Mile', 'Yard'], <Synonyms For Eat>: ['Chow', 'Gobble', 'Scarf', 'Wolf'], <Musicals Beginning With 'C'>: ['Cabaret', 'Carousel', 'Cats', 'Chicago']. Now, given the pool: {}. The answer must be \$*\$ groups, each of them containing 4 words and defined by one category, and the output format must be the same as the example. Give me the answer immediately.

English prompt for 3-group games: \\
I am going to give you a pool of \#*\# words. These \#*\# words can be separated into \$*\$ equal groups of 4 words linked under some category. I want you to tell me the four groups and what category you think connects them. Here is an example: Given the pool ['Water', 'Happiness', 'Fire', 'Earth', 'Mercury', 'Surprise', 'Wind', 'Sadness', 'Venus', 'Pluto', 'Angry', 'Mars'], you would output: <Natural Elements>: ['Water', 'Fire', 'Earth', 'Wind'], <Emotions>: ['Happiness', 'Sadness', 'Angry', 'Surprise'], <Planets>: ['Mercury', 'Venus', 'Pluto', 'Mars']. Now, given the pool: {}. The answer must be \$*\$ groups, each of them containing 4 words and defined by one category, and the output format must be the same as the example. Give me the answer immediately.

English prompt for 4-group games: \\
I am going to give you a pool of \#*\# words. These \#*\# words can be separated into \$*\$ equal groups of 4 words linked under some category. I want you to tell me the four groups and what category you think connects them. Here is an example: Given the pool ['Water', 'Fire', 'Sad', 'Wind', 'Happy', 'Earth', 'Angry', 'Surprised'], you would output: <Natural Elements>: ['Water', 'Fire', 'Earth', 'Wind'], <Emotions>: ['Happy','Sad','Angry','Surprised']. Now, given the pool: {}. The answer must be \$*\$ groups, each of them containing 4 words and defined by one category, and the output format must be the same as the example. Give me the answer immediately.

\subsection{Model Response Reformatting Prompt}
English prompt for model response reformatting: \\
Please reformat the following response into a format mapping topic names, encased by < and > with no single quotes, to a list of Python single-quote strings containing the words in a topic (removing the single quotes within each word, like the ' in 'Newton's'). An example of the format is: <Water>: ['Lake', 'River', 'Pond', 'Ocean'], <Family>: ['Uncle', 'Niece', 'Mom', 'Sister'], <American Food>: ['Burger', 'Pizza', 'Wings', 'Steak'], <Precious Metals>: ['Gold', 'Rhodium', 'Platinum', 'Nickel Silver']. Please reformat the following response, outputting only the final reformatted version:

\subsection{Candidate Groups for Word Overlap Prompt}
English prompt for word overlap grouping proposal for 4-group 4-word games (as other game sizes follow a similar format): \\
I am going to give you a list of words, as well as a list of topics each word could fall under. Please group the words under each topic, with replacement, allowing words to be grouped under several topics. Each word must be used at least once. Output only these groups in Python dictionary format, mapping a string representing the topic to a list of strings representing the selected candidate words for that topics. Here is an example -- given 4 topics ["NBA Teams", "Units of Length", "Synonyms For Eat", "Musicals Beginning With 'C'"] and 16 words ["Bucks", "Foot", "Chow", "Gobble", "Scarf", "Cabaret", "Carousel", "Cats", "Chicago", "Wolf", "League", "Mile", "Yard", "Heat", "Jazz", "Nets"], you could output in Python dictionary format: "NBA TEAMS": ["Bucks", "Heat", "Jazz", "Nets", "Mile", "Chicago"], "UNITS OF LENGTH": ["Foot", "League", "Mile", "Yard", "Carousel", "Bucks"], "Synonyms For Eat": ["Chow", "Gobble", "Scarf", "Wolf", "Cats", "Bucks"], "Musicals Beginning With 'C'": ["Cabaret", "Carousel", "Cats", "Chicago", "Chow"]. Here is the list of topics: LIST\_TOPICS . Here is the pool of words from which you can select from: LIST\_WORDS. Output solely the groups in the previously specified Python dictionary format.

\section{LLM Parameter Settings}
\label{app:parameter}
For the efficiency and accuracy of our evaluation process, we fixed the parameters for the same model during different game evaluations. As our task requires LLMs to have flexible thoughts to explore the internal connections between words and group them, we maintain the default settings for open-source models, as official documents suggest. In addition, we found that greedy sampling can generate nonsense responses with occasional repetitions of one single word. Therefore, we didn’t use it in our open-source tasks.

\begin{itemize}
\item Parameter settings for Llama-3-8B-Instruct and Mistral-7B-Instruct-v0.2:
\begin{itemize}
\item do\_sample: \textit{True}
\item top\_k: \textit{50}
\item top\_p: \textit{0.9}
\item temperature: \textit{0.6}
\item num\_return\_sequences: \textit{1}
\item max\_length: \textit{512}
\end{itemize}

\item Parameter settings for Aya-23-8B:
\begin{itemize}
\item do\_sample: \textit{True}
\item temperature: \textit{0.3}
\item num\_return\_sequences: \textit{1}
\item max\_length: \textit{512}
\end{itemize}
\end{itemize}

\section{Handling the Risk of Multiple Solutions Per Game}
\label{app:multiple-solutions}

When generating a given game, we ensure that there are no repeating words or repeating group topics within a game. This ensures that no more than one solution to the game, utilizing the groupings from which the games are constructed, exists per game.

While there may be additional possible groupings that could be created from a game’s pool (like answering with groupings that do not natively exist in a games’ language’s groupings dataset, such as an additional word counting under a grouping topic, or swapping $N$ words between groups), this possibility is accounted for as an aspect of difficulty in the game through our word overlap analysis. Additional grouping variabilities can also occur in answers, such as more/less words, or repeated words in a grouping; however, these variabilities are due to lapses in reasoning from the LLMs, such as misunderstanding the task or the output format.

Additionally, as we consider the ``correct answer'' for a game to be all correct groupings submitted (while still scoring individual groupings for the sake of evaluation and analysis), each game would require $N$ complete groupings as the correct solution. While there can be individual grouping member variations (such as word overlap or the potential to ``force'' words into groupings with new topics by stretching word meaning or usage), for this to constitute a new solution for a game, this would need to occur across all groupings in the game, which is highly unlikely. While there may be \textit{plausible} solutions to each \ours{} game (as there exist plausible solutions as one plays \textit{Connections}, the inspiration behind \ours{}), these do not constitute complete solutions.

\section{Handling Additional Groups in Model Responses}
\label{app:additional-groups-in-responses}
We considered penalizing the grouping evaluation score if a model provided additional groupings, but however opted instead to only consider $N$ answered groupings per game (where $N$ is the group count for a game) for a few reasons. While there is the potential for a model to output extra groupings for a ``game-breaking'' reason (such as outputting all permutations of groupings for a game to ensure attaining the correct answer), this phenomenon has not been observed, with models outputting extra groupings for non-game-breaking reasons being far more common.

For example, we can observe that models can partially repeat their answer, or following their answer, output smaller straggler groups that do not match the game group size requirements. Here is an example response from Llama-3.1-8B in the \textsc{es-en} 4-group/4-word dataset:
\begin{itemize}
    \item “Here is the solution: The four groups are: <VEGETABLES>: ['Celery', 'Carrot', 'Broccoli', 'Leaf'], <PLANT PARTS>: ['Branch', 'Trunk', 'Root', 'Leaf'], <WRITING TOOLS>: ['Pen', 'Pencil', 'Eraser', 'Envelope'], <VERBS>: ['Sang', 'Leaped', 'Yelled', 'Cried']. The category that connects them is that they are all related to everyday life, but they belong to different domains. The first group is related to food, the second is related to plants, the third is related to writing and communication, and the fourth is related to human actions and emotions. Here is the answer in the required format: <VEGETABLES>: ['Celery', 'Carrot', 'Broccoli', 'Leaf'], <PLANT PARTS>: ['Branch', 'Trunk', 'Root', 'Leaf'], <WRITING TOOLS>: ['Pen',"
\end{itemize}

Additionally, models sometimes restate their groupings in subsequent discussion in their answer, sometimes with small variations (like omitting a group, or omitting a word in a group), or in a ``reasoning process'', as we see models correct themselves in their own answer. Here is an example response from Llama-3.1-8B in the \textsc{es-en} 4-group/4-word dataset, where the model corrects itself (with this correction being used for scoring):
\begin{itemize}
    \item "Solution: The 16 words can be separated into four groups linked under the category of ""Titles of Positions in the Catholic Church"". Here is the output:<TITLES OF POSITIONS IN THE CATHOLIC CHURCH>: ['Archbishop', 'Cardinal', 'Father', 'Priest'], <UNITS OF LENGTH>: ['Cool', 'Ho', 'Powder', 'Shell'], <COOKIES>: ['Shortbread', 'Solid', 'Fantasy', 'Cóc'], <RELIGIOUS TITLES>: ['Man', 'Monk', 'Pas', 'Roll']. Note: The word 'Fantasy' does not belong to the category, it should be 'Fantastic'. Also, the word 'Pas' does not belong to the category, it should be 'Coc'. The correct answer is:<TITLES OF POSITIONS IN THE CATHOLIC CHURCH>: ['Archbishop', 'Cardinal', 'Father', 'Priest'], <UNITS OF","
\end{itemize}

In these situations, penalizing models for their responses having extra groups will then unfairly penalize models that offer additional discussion aside from solely their response (where they restate their groupings and inconsistencies can creep in), or when they provide reasoning steps toward their response (with intermediate groupings then being penalized). Because of the frequency of these cases observed in preliminary testing, and as observed in our evaluations using \ours{}, we opt instead not to penalize models for providing extra groupings, instead only considering $N$ groupings per model response.

\section{Determining Topic Achieved}
\label{app:determining-ta}
The threshold value of 0.3 was selected through experimentation. The basis of the Topic Achieved score was evaluated using both BERTScore and FastText embeddings-based cosine similarity as measures of similarity between topics. 7 Different threshold values were used for score generation (0.1, 0.2, 0.3, 0.4, 0.5, 0.6, and 0.7). For 100 games of the 2-group, 2-word \textsc{zh} dataset, we had human annotations of topic achieved scores. Human annotators were provided the same rules for calculating the Topic Achieved score but were asked to base it on their own perception of similarity. As a means of selecting the metric most similar to human judgment, we calculated the inner-annotator agreement utilizing Randolph's Kappa between the human annotations and each of the methods used for calculation. From this, we found that a FastText embeddings-based approach with a similarity threshold of 0.3 performed best, attaining a Randolph's Kappa score of 0.55. A BERTScore-based approach with a similarity threshold of 0.5 had the second-best score of 0.53.

Additionally, due to the lack of context provided in \textit{Connections} and \ours{} to inform word meaning and group membership, semantic embeddings were preferred.

\section{Annotator Information}
\label{app:annotator-information}
\subsection{Annotator Request}
Annotators were initially provided with the following message, choosing to submit groupings for usage on a volunteer basis:

"Hello [ANNOTATOR NAME]! My name is [NAME], and I am an working with [NAME] on a project using word grouping games (similar to "Connections" from the New York Times) to assess multilingual reasoning. We are creating word grouping datasets in source languages, composed of groupings of four words or very small phrases connected by an associated topic. These topics don't necessarily pertain just to semantic meaning of the words, and can relate to other potential uses/meanings of the words. Here's a sample from our English dataset:

[Teacher, Scientist, Engineer, Doctor] -- Professions
[Pool, Jacuzzi, Ocean, Lake] -- Places to Swim
[Rocky, Appalachian, Adirondack, Sierra Nevada] -- Mountain Ranges
[Hansel, Witch, Bear, Wolf] -- Fairy Tale Figures

The groupings can be connected by non-culture-related topics (i.e., topics whose contents can be recognized across cultures, such as "Professions" or "Places to Swim") or culture-related topics (i.e., topics whose contents are better recognized in cultures associated with that language, such as Western/North American culture being familiar with the specific mountain range names in "Mountain Ranges"). With your knowledge of [LANGUAGE], I wanted to ask if you would be available to write custom word groupings in [LANGUAGE]. We want to collect 25-50 groupings each in additional languages, with an even split between non-culture-related group topics and culture-related group topics per language. No worries at all if that is a bit too much -- however many groupings/languages you could commit to would be fine. I have attached the dataset of our English groupings for your reference. If this would be possible, please let me know! I am available via Slack and via email [EMAIL] if you have any questions."

Annotators, upon responding favorably to this request, were then provided with further information on the task, including how the groups would be used (random sampling to create games) and providing examples when needed. Annotators were then informed of the translation protocol for translating groupings in their source language into English. This is further detailed in Appendix \ref{app:translations-protocol}.

\subsection{Annotator Demographics}
The 2 annotators for the \textsc{en} groupings had the following background statistics:
\begin{itemize}
\item 2 were university-educated
\item 2 were born and raised in the United States
\item 1 was male, 1 was female
\end{itemize}

The 7 annotators for the \textsc{es} groupings had the following background statistics:
\begin{itemize}
\item 5 were university-educated, 2 were not
\item 3 were born and raised in Mexico, 4 were born and raised in the United States within Mexican families
\item 3 were male, 4 were female
\end{itemize}

The 4 annotators for the \textsc{zh} groupings had the following background statistics:
\begin{itemize}
\item 2 were university-educated, 2 were not
\item 4 were all born and raised in China.
\item 1 were male, 3 were female
\end{itemize}

In addition, some Chinese groupings were created by searching and checking online resources, as well as Chinese LLMs such as the Baidu ERNIE Bot (\url{https://yiyan.baidu.com/}). We used Baidu ERNIE for several simple Chinese groups creation. For example, we used the prompt: "Please list several animals that are common to see in our daily lives". We then manually checked and picked 4 animals in Chinese from their answers to form one simple Chinese group, for example: cat, dog, bird, and fish.

The 3 annotators for the \textsc{hi} groupings had the following background statistics:
\begin{itemize}
\item 3 were university-educated
\item 2 were born and raised in India, 1 was born in the United Kingdom and raised in the United States
\item 3 were male
\end{itemize}

The 2 annotators for the \textsc{ar} groupings had the following background statistics:
\begin{itemize}
\item 2 were university-educated
\item 2 were born and raised in Egypt
\item 1 was male, 1 was female
\end{itemize}

\subsection{Protocol for Translation into English}
\label{app:translations-protocol}
Annotators were instructed to translate each word within a group, along with the group topic, from the initial language into English. In situations where a direct translation did not exist, annotators were instructed to translate literally, word for word. This process was assisted with the online translation tools Google Translate\footnote{\url{https://translate.google.com/}} and DeepL\footnote{\url{https://www.deepl.com/en/translator}}. The authors verified translations using these tools as well. The same translation process was applied to every grouping within our dataset subsets in order to ensure parity.

\section{Example WGG Games}
\label{app:example-games}
\subsection{en-es}
\begin{table}[H]
\tiny
\begin{tabular}{ccccc}
\hline
\multicolumn{5}{c}{\textsc{en-es} 4-group, 4-word}                                                                                                                                                                                                                         \\ \hline
\multicolumn{4}{c|}{Shuffled Game Word Pool}                                                                                                                 & Topic                                                                                              \\ \hline
{\color[HTML]{3166FF} Neruda}  & {\color[HTML]{6200C9} History} & {\color[HTML]{009901} Ladybug}   & \multicolumn{1}{c|}{{\color[HTML]{009901} Mosquito}}    & {\color[HTML]{009901} Bugs}                                                                        \\
{\color[HTML]{6200C9} Mystery} & {\color[HTML]{6200C9} Poems}   & {\color[HTML]{3166FF} Mistral}   & \multicolumn{1}{c|}{{\color[HTML]{9A0000} Chilindrina}} & {\color[HTML]{3166FF} \begin{tabular}[c]{@{}c@{}}Latin-American\\ Poets\end{tabular}}              \\
{\color[HTML]{009901} Locust}  & {\color[HTML]{9A0000} Chavo}   & {\color[HTML]{009901} Butterfly} & \multicolumn{1}{c|}{{\color[HTML]{3166FF} Martí}}       & {\color[HTML]{6200C9} Genres}                                                                      \\
{\color[HTML]{9A0000} Ramón}   & {\color[HTML]{3166FF} Acevedo} & {\color[HTML]{9A0000} Quico}     & \multicolumn{1}{c|}{{\color[HTML]{6200C9} Romance}}     & {\color[HTML]{9A0000} \begin{tabular}[c]{@{}c@{}}Characters from\\ El Chavo del Ocho\end{tabular}} \\ \hline
\end{tabular}
\caption{An example 4-group, 4-word game from the \textsc{en-es} dataset background. Here, you can see a mix of both colloquial topics, such as "Latin-American Poets", and non-colloquial topics, such as "Bugs".}
\label{tab:example-en-es-4g4w}
\end{table}
\subsection{en-zh}
\begin{table}[H]
\tiny
\begin{tabular}{ccccc}
\hline
\multicolumn{5}{c}{\textsc{en-zh} 4-group, 4-word}                                                                                                                                                                                                                                                                                                                                                                                          \\ \hline
\multicolumn{4}{c|}{Shuffled Game Word Pool}                                                                                                                                                                                                                                                                                              & Topic                                                                                  \\ \hline
{\color[HTML]{009901} Positivism}                                                        & {\color[HTML]{009901} Idealism}                                                 & {\color[HTML]{9A0000} Great Wall}     & \multicolumn{1}{c|}{{\color[HTML]{9A0000} \begin{tabular}[c]{@{}c@{}}Terracotta \\ Warriors\\ and Horses\end{tabular}}} & {\color[HTML]{3166FF} Ruminants}                                                       \\
{\color[HTML]{3166FF} Ox}                                                                & {\color[HTML]{9A0000} \begin{tabular}[c]{@{}c@{}}Forbidden\\ City\end{tabular}} & {\color[HTML]{009901} Existentialism} & \multicolumn{1}{c|}{{\color[HTML]{6200C9} Water}}                                                                    & {\color[HTML]{9A0000} \begin{tabular}[c]{@{}c@{}}Ancient\\ Building\end{tabular}}      \\
{\color[HTML]{9A0000} \begin{tabular}[c]{@{}c@{}}Qin Shi Huang\\ Mausoleum\end{tabular}} & {\color[HTML]{6200C9} Fire}                                                     & {\color[HTML]{6200C9} Earth}          & \multicolumn{1}{c|}{{\color[HTML]{3166FF} Sheep}}                                                                    & {\color[HTML]{6200C9} \begin{tabular}[c]{@{}c@{}}Natural\\ Elements\end{tabular}}      \\
{\color[HTML]{3166FF} Reindeer}                                                          & {\color[HTML]{009901} Pragmatism}                                               & {\color[HTML]{6200C9} Wind}           & \multicolumn{1}{c|}{{\color[HTML]{3166FF} Sika Deer}}                                                                & {\color[HTML]{009901} \begin{tabular}[c]{@{}c@{}}Philosophical\\ Concept\end{tabular}} \\ \hline
\end{tabular}
\caption{An example 4-group, 4-word game from the \textsc{en-zh} dataset background. Here, you can see a mix of both colloquial topics, such as "Ancient Building", and non-colloquial topics, such as "Natural Elements".}
\label{tab:example-en-zh-4g4w}
\end{table}

\section{Group Evaluation Metric Comparison}
\label{app:group-evaluation-metrics}
\begin{figure}[h]
    \centering
    \includegraphics[width=\columnwidth]{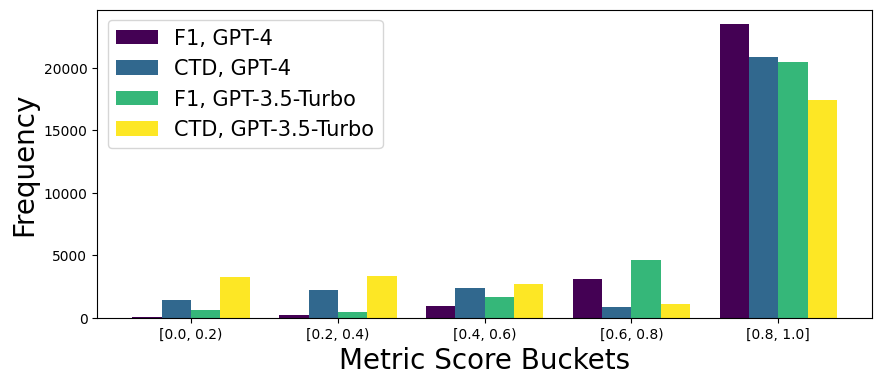}
    \caption{Depicting the distribution of game average F1 and CTD scores across 5 buckets of score ranges, encapsulating the total score range for both metrics. This is evaluated for closed-source models.}
    \label{fig_grouping_score_distribution}
\end{figure}
As can be seen in Figure \ref{fig_grouping_score_distribution}, CTD invokes a great penalization on model responses, with peaks in lower-valued score buckets and a sharp decrease in median-valued buckets suggesting a lack of granularity in model response scoring. Contrastingly, F1 score offers a gradual increase in frequency in increasing-valued score buckets, and no peaks at opposing ends of the score value range, providing a more complete and fine-toothed scoring of model performance in this task.

\section{Additional Model Performance Analyses}
\label{app:section-additional-analyses}

\subsection{Performance Relative to Topical Tagging of \ours{} Groupings}
\begin{table}[]
\centering
\tiny
\begin{tabular}{@{}cccc@{}}
\toprule
\textbf{Models}                         & \multicolumn{3}{c}{\textbf{\begin{tabular}[c]{@{}c@{}}Mean F1 Score in\\ Groups of Tag\end{tabular}}}                                                     \\ \midrule
\multicolumn{1}{c|}{\textit{Group Tag}} & \begin{tabular}[c]{@{}c@{}}General\\ Knowledge\end{tabular} & \begin{tabular}[c]{@{}c@{}}Cultural and Pop\\ Culture Knowledge\end{tabular} & Linguistic \\ \midrule
\multicolumn{1}{c|}{GPT-3.5-Turbo}      & \textbf{0.902}                                              & 0.871                                                                          & 0.826      \\
\multicolumn{1}{c|}{GPT-4}              & \textbf{0.954}                                              & 0.937                                                                          & 0.916      \\ \midrule
\multicolumn{1}{c|}{Llama3-8B}          & \textbf{0.804}                                              & 0.767                                                                          & 0.724      \\
\multicolumn{1}{c|}{Llama3.1-70B}       & \textbf{0.935}                                              & 0.900                                                                          & 0.859      \\
\multicolumn{1}{c|}{Mistral-7B}         & \textbf{0.779}                                              & 0.694                                                                          & 0.669      \\
\multicolumn{1}{c|}{Aya-8B}             & \textbf{0.866}                                              & 0.818                                                                          & 0.789      \\ \bottomrule
\end{tabular}
\caption{The averaged performance on groups, separated by tag, by model across each non-NYT 4-group, 4-word game used during experimentation. Results are divided by each group tag and between closed- and open-source models.}
\label{tab:4g4w-tag-results}
\end{table}
\begin{table*}[]
\centering
\tiny
\begin{tabular}{@{}ccccllllllllllll@{}}
\toprule
\textbf{Models}                         & \multicolumn{15}{c}{\textbf{\begin{tabular}[c]{@{}c@{}}Mean F1 Score in\\ Groups of Tag, By Dataset\end{tabular}}}                                                                                                                                                                                                                                                                                                                                    \\ \midrule
\multicolumn{1}{c|}{\textit{Dataset}}   & \multicolumn{3}{c|}{es}                                                                              & \multicolumn{3}{c|}{es-en}                                                              & \multicolumn{3}{c|}{zh}                                                        & \multicolumn{3}{c|}{zh-en}                                                     & \multicolumn{3}{c|}{}                                                              \\
\multicolumn{1}{c|}{\textit{Group Tag}} & GK                                 & CPCK                      & \multicolumn{1}{c|}{L}              & \multicolumn{1}{c}{GK} & \multicolumn{1}{c}{CPCK} & \multicolumn{1}{c|}{L}              & \multicolumn{1}{c}{GK} & \multicolumn{1}{c}{CPCK} & \multicolumn{1}{c|}{L}     & \multicolumn{1}{c}{GK} & \multicolumn{1}{c}{CPCK} & \multicolumn{1}{c|}{L}     & \multicolumn{1}{c}{}   & \multicolumn{1}{c}{}     & \multicolumn{1}{c|}{}          \\ \midrule
\multicolumn{1}{c|}{GPT-3.5-Turbo}      & \textbf{0.909}                     & 0.810                     & \multicolumn{1}{c|}{0.774}          & \textbf{0.949}         & 0.841                    & \multicolumn{1}{l|}{0.781}          & \textbf{0.959}         & 0.907                    & \multicolumn{1}{l|}{0.904} & 0.841                  & \textbf{0.866}           & \multicolumn{1}{l|}{0.761} & \textbf{}              &                          & \multicolumn{1}{l|}{}          \\
\multicolumn{1}{c|}{GPT-4}              & \textbf{0.950}                     & 0.890                     & \multicolumn{1}{c|}{0.840}          & \textbf{0.977}         & 0.924                    & \multicolumn{1}{l|}{0.856}          & \textbf{0.981}         & 0.965                    & \multicolumn{1}{l|}{0.960} & 0.932                  & \textbf{0.954}           & \multicolumn{1}{l|}{0.900} &                        &                          & \multicolumn{1}{l|}{\textbf{}} \\ \midrule
\multicolumn{1}{c|}{Llama3-8B}          & \textbf{0.786}                     & 0.656                     & \multicolumn{1}{c|}{0.600}          & \textbf{0.833}         & 0.723                    & \multicolumn{1}{l|}{0.680}          & 0.635                  & \textbf{0.649}           & \multicolumn{1}{l|}{0.585} & 0.845                  & \textbf{0.882}           & \multicolumn{1}{l|}{0.788} & \textbf{}              &                          & \multicolumn{1}{l|}{}          \\
\multicolumn{1}{c|}{Llama3.1-70B}       & \textbf{0.894}                     & 0.831                     & \multicolumn{1}{c|}{0.766}          & \textbf{0.941}         & 0.857                    & \multicolumn{1}{l|}{0.783}          & \textbf{0.982}         & 0.948                    & \multicolumn{1}{l|}{0.906} & \textbf{0.964}         & 0.946                    & \multicolumn{1}{l|}{0.899} & \textbf{}              &                          & \multicolumn{1}{l|}{}          \\
\multicolumn{1}{c|}{Mistral-7B}         & \textbf{0.705}                     & 0.536                     & \multicolumn{1}{c|}{0.470}          & \textbf{0.809}         & 0.644                    & \multicolumn{1}{l|}{0.589}          & \textbf{0.820}         & 0.746                    & \multicolumn{1}{l|}{0.719} & \textbf{0.857}         & 0.846                    & \multicolumn{1}{l|}{0.718} & \textbf{}              &                          & \multicolumn{1}{l|}{}          \\
\multicolumn{1}{c|}{Aya-8B}             & \textbf{0.816}                     & 0.722                     & \multicolumn{1}{c|}{0.707}          & \textbf{0.881}         & 0.754                    & \multicolumn{1}{l|}{0.759}          & \textbf{0.930}         & 0.881                    & \multicolumn{1}{l|}{0.783} & 0.859                  & \textbf{0.863}           & \multicolumn{1}{l|}{0.770} & \textbf{}              &                          & \multicolumn{1}{l|}{}          \\ \midrule
\multicolumn{1}{c|}{\textit{Dataset}}   & \multicolumn{3}{c|}{hi}                                                                              & \multicolumn{3}{c|}{hi-en}                                                              & \multicolumn{3}{c|}{ar}                                                        & \multicolumn{3}{c|}{ar-en}                                                     & \multicolumn{3}{c|}{en}                                                            \\
\multicolumn{1}{c|}{\textit{Group Tag}} & GK                                 & CPCK                      & \multicolumn{1}{c|}{L}              & \multicolumn{1}{c}{GK} & \multicolumn{1}{c}{CPCK} & \multicolumn{1}{c|}{L}              & \multicolumn{1}{c}{GK} & \multicolumn{1}{c}{CPCK} & \multicolumn{1}{c|}{L}     & \multicolumn{1}{c}{GK} & \multicolumn{1}{c}{CPCK} & \multicolumn{1}{c|}{L}     & \multicolumn{1}{c}{GK} & \multicolumn{1}{c}{CPCK} & \multicolumn{1}{c|}{L}         \\ \midrule
\multicolumn{1}{c|}{GPT-3.5-Turbo}      & \multicolumn{1}{l}{\textbf{0.898}} & \multicolumn{1}{l}{0.882} & \multicolumn{1}{l|}{0.851}          & \textbf{0.957}         & 0.940                    & \multicolumn{1}{l|}{0.952}          & \textbf{0.779}         & 0.765                    & \multicolumn{1}{l|}{0.702} & \textbf{0.933}         & 0.911                    & \multicolumn{1}{l|}{0.895} & 0.889                  & \textbf{0.916}           & \multicolumn{1}{l|}{0.811}     \\
\multicolumn{1}{c|}{GPT-4}              & \multicolumn{1}{l}{0.963}          & \multicolumn{1}{l}{0.960} & \multicolumn{1}{l|}{\textbf{0.979}} & 0.978                  & 0.951                    & \multicolumn{1}{l|}{\textbf{0.984}} & 0.882                  & \textbf{0.888}           & \multicolumn{1}{l|}{0.852} & \textbf{0.968}         & 0.944                    & \multicolumn{1}{l|}{0.957} & \textbf{0.959}         & 0.956                    & \multicolumn{1}{l|}{0.911}     \\ \midrule
\multicolumn{1}{c|}{Llama3-8B}          & \multicolumn{1}{l}{\textbf{0.766}} & \multicolumn{1}{l}{0.703} & \multicolumn{1}{l|}{0.665}          & 0.879                  & \textbf{0.880}           & \multicolumn{1}{l|}{0.865}          & 0.830                  & \textbf{0.838}           & \multicolumn{1}{l|}{0.814} & \textbf{0.859}         & 0.775                    & \multicolumn{1}{l|}{0.799} & 0.800                  & \textbf{0.803}           & \multicolumn{1}{l|}{0.719}     \\
\multicolumn{1}{c|}{Llama3.1-70B}       & \multicolumn{1}{l}{\textbf{0.944}} & \multicolumn{1}{l}{0.943} & \multicolumn{1}{l|}{0.927}          & \textbf{0.968}         & 0.947                    & \multicolumn{1}{l|}{\textbf{0.968}} & \textbf{0.860}         & 0.823                    & \multicolumn{1}{l|}{0.759} & \textbf{0.945}         & 0.900                    & \multicolumn{1}{l|}{0.888} & \textbf{0.915}         & 0.909                    & \multicolumn{1}{l|}{0.832}     \\
\multicolumn{1}{c|}{Mistral-7B}         & \multicolumn{1}{l}{\textbf{0.545}} & \multicolumn{1}{l}{0.464} & \multicolumn{1}{l|}{0.474}          & \textbf{0.941}         & 0.915                    & \multicolumn{1}{l|}{0.939}          & \textbf{0.652}         & 0.577                    & \multicolumn{1}{l|}{0.588} & \textbf{0.850}         & 0.728                    & \multicolumn{1}{l|}{0.771} & \textbf{0.833}         & 0.789                    & \multicolumn{1}{l|}{0.749}     \\
\multicolumn{1}{c|}{Aya-8B}             & \multicolumn{1}{l}{\textbf{0.914}} & \multicolumn{1}{l}{0.846} & \multicolumn{1}{l|}{0.858}          & \textbf{0.958}         & 0.928                    & \multicolumn{1}{l|}{0.931}          & 0.713                  & \textbf{0.760}           & \multicolumn{1}{l|}{0.736} & \textbf{0.864}         & 0.774                    & \multicolumn{1}{l|}{0.787} & \textbf{0.856}         & 0.837                    & \multicolumn{1}{l|}{0.769}     \\ \bottomrule
\end{tabular}

\caption{The averaged performance on groups, separated by tag, by model across each non-NYT 4-group, 4-word game used during experimentation. Results are divided by each dataset, by each group tag, and between closed- and open-source models. GK = General Knowledge, CPCK = Cultural and Pop Culture Knowledge, L = Linguistic}
\label{tab:4g4w-tag-by-dataset-results}
\end{table*}

We additionally perform tagging/typing of each grouping in our grouping datasets, categorizing groupings under three labels:
\begin{itemize}
    \item \textbf{General Knowledge}, where groupings pertain to encyclopaedic knowledge or knowledge gained from everyday life, largely agnostic of cultural origin
    \item \textbf{Cultural and Pop Culture Knowledge}, where groupings pertain to knowledge strongly grounded in culture (such as food, religion, important sites), or pop culture (such as movies, music, and sports)
    \item \textbf{Linguistic}, where groupings pertain to linguistic concepts, such as synonyms, morphology, phonetics, idioms, and slang
\end{itemize}
We do not further type groupings within each category, such as “Linguistic”, due to a small number of examples per category.

This annotation was performed by two annotators (university-educated, male, Indian and American backgrounds) until full agreement was reached. From this, we perform an analysis, looking at F1 performance in groups within these tags, across datasets, across models. We report results averaged across all 4-group/4-word datasets, per model, in Table~\ref{tab:4g4w-tag-results}, and separated by language dataset in Table~\ref{tab:4g4w-tag-by-dataset-results}.

From this, we can observe a clear trend in performance relative to the type of content present within a grouping – across all models, average performance in groupings tagged as a category averaged across all datasets decreases by $~20\%$ from “General Knowledge” to “Cultural and Pop Culture Knowledge” to “Linguistic”. We can observe similar trends with the results split by the respective language dataset, as in Table~\ref{tab:4g4w-tag-by-dataset-results}.

\subsection{Performance with \textit{Connections} Games}
\label{app:connection-difficulty}
\begin{table}[tbp]
\centering
\tiny
\begin{tabular}{@{}l|cccc@{}}
\toprule
\multicolumn{1}{c|}{\textbf{Models}}      & \multicolumn{4}{c}{\textbf{F1 Score}}                                     \\ \midrule
\textit{Color Difficulty}                 & Yellow              & Green               & Blue          & Purple        \\
\multicolumn{1}{c|}{\textit{Backgrounds}} & nyt shuf.           & nyt shuf.           & nyt shuf.     & nyt shuf.     \\ \midrule
GPT-3.5-Turbo                             & \textbf{0.671}      & 0.638               & 0.575         & 0.523         \\
\multicolumn{1}{c|}{GPT-4}                & \textbf{0.904}      & 0.876               & 0.805         & 0.702         \\ \midrule
                                          & \multicolumn{4}{c}{\textbf{\% FastText Topic Achieved (threshold = 0.3)}} \\ \midrule
GPT-3.5-Turbo                             & \textbf{0.365}      & 0.303               & 0.337         & 0.259         \\
\multicolumn{1}{c|}{GPT-4}                & 0.508               & \textbf{0.525}      & 0.507         & 0.365         \\ \bottomrule
\end{tabular}
\caption{The averaged results on \textit{Connections} groupings for closed-source models across each 4-group, 4-word nyt shuf. game used during experimentation. Results are divided by the \textit{Connections} color difficulty associated with a given grouping. The color difficulties are ordered yellow, green, blue, purple, from easiest to hardest.}
\label{tab:nyt-color-4g4w-results}
\end{table}
One can note in Table~\ref{tab:4g4w-raw-results} how performance greatly differs between the custom groupings-sourced games and the \textit{Connections} groupings-sourced games. However, it is important to note the difficulty paradigm present within \textit{Connections} -- each game features groupings of four different difficulties, represented by the assigned color of a grouping, with difficulty said to be scaled from easiest to hardest in the order of yellow, green, blue, and purple. The \textsc{nyt seq.} and \textsc{nyt shuf.} games utilized in Table~\ref{tab:4g4w-raw-results} utilize groupings of all difficulties. When stratifying performance in \textsc{nyt shuf.} (chosen to act as a better comparison to the randomly-generated custom groupings-sourced games) by the difficulty color, as seen in Table~\ref{tab:nyt-color-4g4w-results}, we can see clear trends in F1 score relative to the color difficulty. This aligns with the listed difficulty scaling of these colors, with F1 score decreasing from yellow, to green, to blue, to purple. This reaffirms the listed difficulties of these colors in the \textit{Connections} games, with difficulty (represented by a decrease in performance) following the prescribed pattern.

We can utilize performance on our \textsc{en} games to compare our efforts in grouping creation to the groupings seen in \textit{Connections}. By comparing performance in \textsc{en} from Table~\ref{tab:4g4w-raw-results} to performance stratified by difficulty color in Table~\ref{tab:nyt-color-4g4w-results}, we can see the closest similarity to yellow groupings from \textit{Connections}. This aligns with our intention to create an easier, more discriminative benchmark, as discussed in Section \ref{sec:related work}.

\subsection{Evaluating True and Predicted Groups as Clusters with Silhouette Score}
\label{app:groups-as-clusters}
Additional insight can be gained by evaluating the true and predicted groupings of a game as if they were clusters. By using unsupervised cluster metrics, such as silhouette score, we can assess the inter-word similarity of a true/guessed grouping relative to the other groupings of the same classification in the same game. This allows us to then see comparisons in trends of silhouette scores for true and predicted groupings, offering a perspective into a possible basis for model group formation, and how that compares to actual groupings in a game. A silhouette score of 0 would mean that word groupings are just as dissimilar within-group as they are between-group, meaning that groupings are not solely composed by semantically similar words, as represented by their word embeddings.

For each scored game, we evaluate the silhouette score of the set of true groupings and the set of guessed groupings. We compile this by ranges of silhouette scores, reporting the percentage of total games observed to have a silhouette score within a given value bucket for each classification of grouping. By comparing these distributions, we can see if models tend to reason along the intended abstract line of reasoning for a given game (i.e. grouping words based on abstract connection, represented by 0-centered silhouette scores), rather than purely by word similarity (represented by silhouette scores at the ends of the metric range). In order to investigate potential differences in this aspect relative to performance, pointing towards different group formation paradigms leading to increased or decreased performance, we stratify these results by the F1 score achieved in a game.

\begin{figure}
    \centering
    \includegraphics[width=\columnwidth]{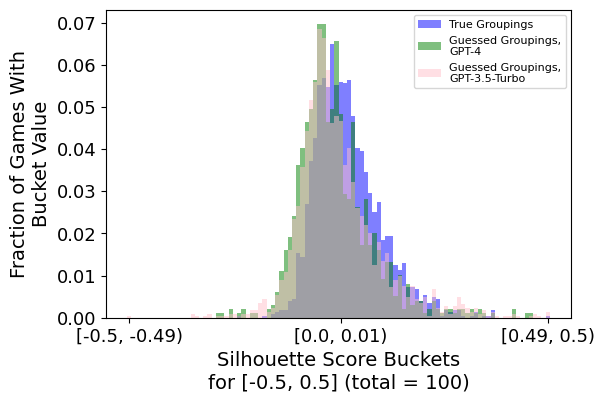}
    \caption{Depicting the distribution of total game composition of games with silhouette score values across 100 buckets of silhouette score ranges, for games where a model scored an F1 $\le$ 0.5. This is evaluated for closed-source models, with game silhouette scores calculated using both the true groupings, and guessed groupings for a given game.}
    \label{fig_silhouette_score_distribution_f1lte05}
\end{figure}
\begin{figure}
    \centering
    \includegraphics[width=\columnwidth]{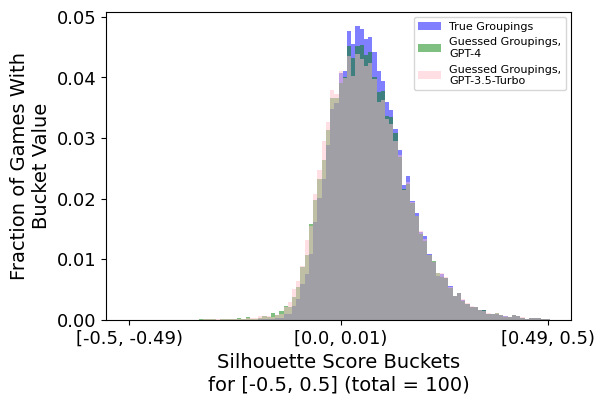}
    \caption{Depicting the distribution of total game composition of games with silhouette score values across 100 buckets of silhouette score ranges, for games where a model scored an F1 $>$ 0.5. This is evaluated for closed-source models, with game silhouette scores calculated using both the true groupings, and guessed groupings for a given game.}
    \label{fig_silhouette_score_distribution_f1gt05}
\end{figure}

Due to the overlapping distributions for the varying classifications of groupings across the evaluated models, as seen in Figures \ref{fig_silhouette_score_distribution_f1lte05} and \ref{fig_silhouette_score_distribution_f1gt05}, we can infer that models employ reasoning patterns reminiscent of the intended abstract line of reasoning for WGG games. This is further emphasized by these distributions trending away from the extremes of the silhouette score ranges, being 0-centered, showing that a lack of inter-word similarity is common for both true and predicted groups, as predicted.

\subsection{Additional Raw Result Analysis}
\label{fig_4g4w_raw_results_bar_plot}
\begin{figure}
    \centering
    \includegraphics[width=\columnwidth]{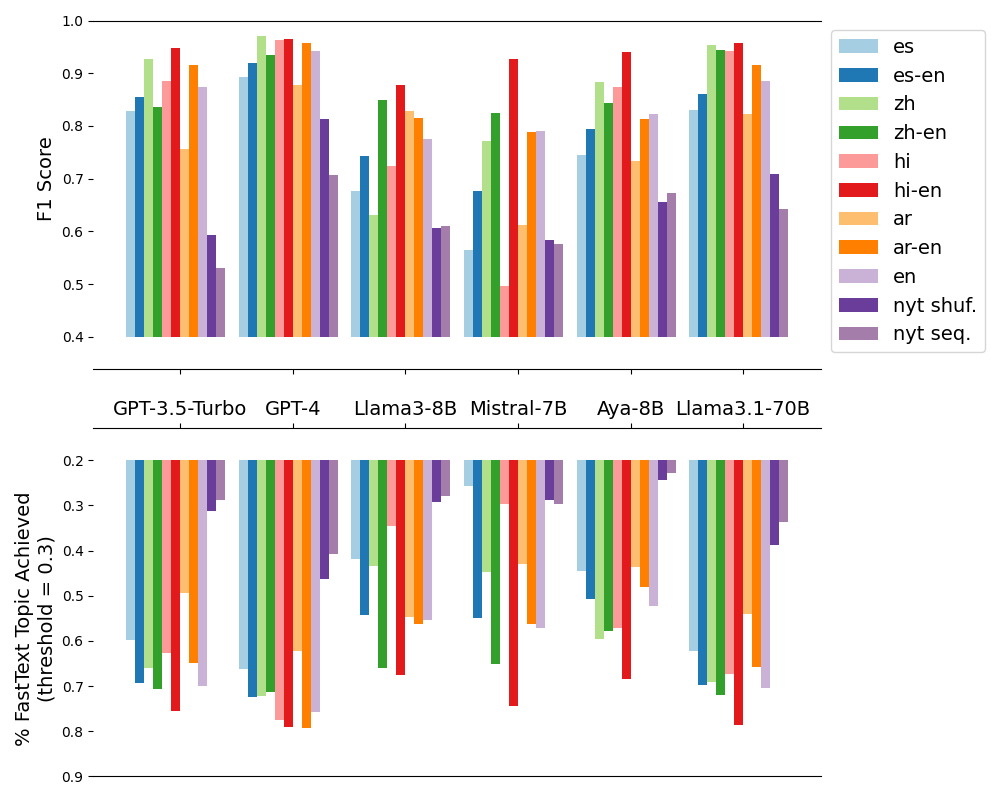}
    \caption{A figure depicting the averaged results by model across each 4-group, 4-word game used during experimentation.}
    \label{fig:f1_raw_results_bar_plot}
\end{figure}

\label{app:cultural_relatedness_difference_by_dataset}
\begin{table*}[]
\tiny
\centering
\begin{tabular}{@{}cccccllllcl@{}}
\toprule
\textbf{Models}                        & \multicolumn{9}{c}{\textbf{Mean F1 Score Difference (Non-Culturally-Related - Culturally-Related)}}                                                                                                                                                          &                                    \\ \midrule
\multicolumn{1}{c|}{\textit{Datasets}} & es             & es-en          & zh             & zh-en          & \multicolumn{1}{c}{hi}             & \multicolumn{1}{c}{hi-en}          & \multicolumn{1}{c}{ar}             & \multicolumn{1}{c}{ar-en}          & \multicolumn{1}{c|}{en}              & Mean                               \\ \midrule
\multicolumn{1}{c|}{GPT-3.5-Turbo}     & \textbf{0.128} & \textbf{0.130} & \textbf{0.054} & \textbf{0.041} & \multicolumn{1}{c}{\textbf{0.026}} & \multicolumn{1}{c}{0.024}          & \multicolumn{1}{c}{\textbf{0.054}} & \multicolumn{1}{c}{\textbf{0.036}} & \multicolumn{1}{c|}{-0.047}          & \multicolumn{1}{c}{\textbf{0.050}} \\
\multicolumn{1}{c|}{GPT-4}             & 0.098          & 0.092          & 0.013          & 0.018          & \multicolumn{1}{c}{0.001}          & \multicolumn{1}{c}{\textbf{0.026}} & \multicolumn{1}{c}{0.036}          & \multicolumn{1}{c}{0.022}          & \multicolumn{1}{c|}{\textbf{-0.017}} & \multicolumn{1}{c}{0.032}          \\ \midrule
\multicolumn{1}{c|}{Llama3-8B}         & 0.137          & 0.128          & -0.005         & 0.025          & 0.048                              & -0.003                             & 0.018                              & 0.082                              & \multicolumn{1}{c|}{-0.033}          & 0.044                              \\
\multicolumn{1}{c|}{Llama3.1-70B}      & 0.100          & 0.118          & 0.031          & 0.028          & 0.017                              & 0.019                              & 0.078                              & 0.044                              & \multicolumn{1}{c|}{-0.027}          & 0.045                              \\
\multicolumn{1}{c|}{Mistral-7B}        & \textbf{0.182} & \textbf{0.198} & 0.068          & \textbf{0.061} & 0.042                              & \textbf{0.036}                     & \textbf{0.080}                     & \textbf{0.109}                     & \multicolumn{1}{c|}{-0.005}          & \textbf{0.086}                     \\
\multicolumn{1}{c|}{Aya-8B}            & 0.130          & 0.144          & \textbf{0.070} & 0.052          & \textbf{0.065}                     & 0.032                              & -0.011                             & 0.095                              & \multicolumn{1}{c|}{\textbf{0.008}}  & 0.065                              \\ \bottomrule
\end{tabular}
\caption{A table depicting the average difference in performance in groups labeled as non-culturally-related versus culturally-related in $4\times4$ games per dataset subset for open- and closed-source models. Note the strong pattern of performance improvement with non-culturally-related groupings, across nearly all languages.}
\label{tab:mean-difference-cultural-groups-by-dataset}
\end{table*}

\label{app:translation_performance_difference_figure}
\begin{figure}
    \centering
    \includegraphics[width=\columnwidth]{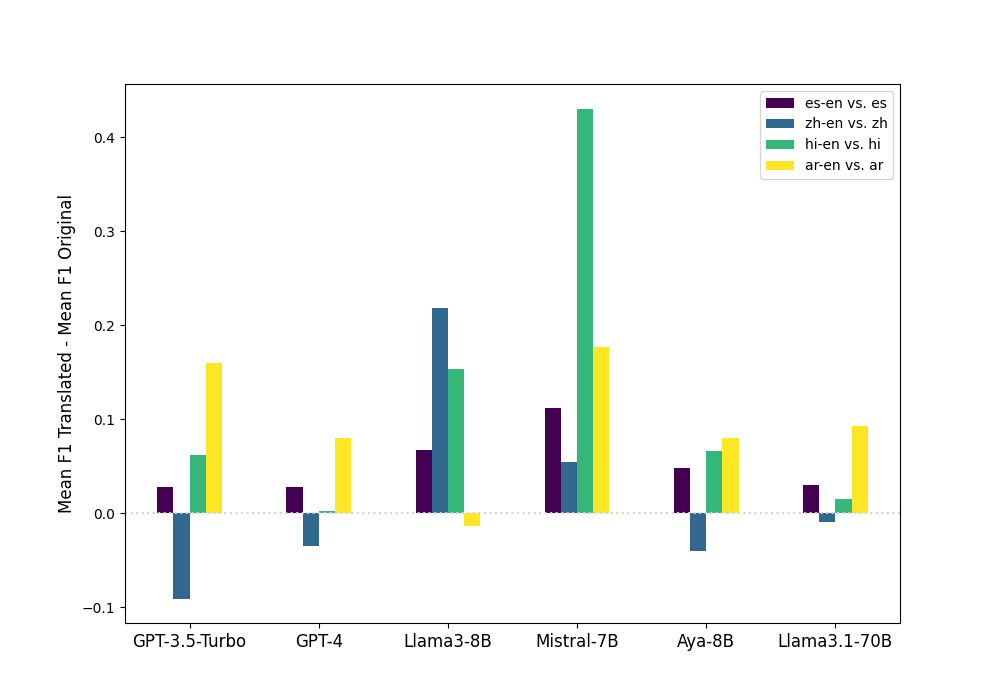}
    \caption{A figure depicting the difference in the average $4\times4$ game results for the translated and non-translated version of each dataset. Note a strong trend of improvement after translation into English, present across nearly all models in nearly all dataset translation pairs.}
    \label{fig:f1_difference_due_to_translation}
\end{figure}

\section{Licenses and Intended Use}
We utilized the OpenAI API to prompt the GPT-3.5-Turbo, GPT-4, and GPT-4o-mini models during evaluation, and for help with reformatting other model responses. This usage complied with OpenAI's terms of use.

We utilized the Llama3-8B and Llama3.1-70B models during evaluation. Our usage complied with Meta's LLAMA 3.1 Community License Agreement.

We utilized the Mistral-7B model during evaluation. Our usage complied with the Apache 2.0 License under which this model was released.

We utilized the Aya-8B model during evaluation. Our usage complied with the Apache 2.0 License under which this model was released.

We utilized pretrained word embeddings from FastText, an embedding library released by Meta, for our various clustering analysis. Our usage complied with the MIT License under which these embeddings were released.

We additionally use packages gensim~\cite{rehurek_lrec}, NLTK~\cite{bird2009natural}, and jieba\footnote{https://github.com/fxsjy/jieba} for text tokenization. We use scikit-learn~\cite{sklearn} for the implementation of KMeans clustering, silhouette score calculation, and adjusted Rand index calculation. We use statsmodels~\cite{seabold2010statsmodels} for the Randolph's Kappa calculation.
\section{Additional Difficulty Metric Analysis}
\label{app:additional_difficulty_metric_analysis}
\subsection{Ranking Correlation for Difficulty Metric Results}
We additionally calculate Spearman's rank correlation coefficient between the difficulty measurements and model performance results for different slices of the benchmark. The categorization of the dataset depends on the chosen difficulty measure, such as varying group sizes or numbers of groups in the game. For continuous difficulty measures, such as clustering metrics based on the Adjusted Rand Index or word overlap, we bin them into three or four categories and calculate the average model performance for each bin. Note that Spearman correlation focuses on ranking order rather than the correlation between the values, so although the scales of these metrics may differ, we argue that variations in their rankings should correspond with variations in model performance rankings. For 10 subsets of the benchmark, i.e., \textsc{en}, \textsc{es}, \textsc{es-en}, \textsc{zh}, \textsc{zh-en}, \textsc{hi}, \textsc{hi-en}, \textsc{ar}, \textsc{ar-en}, and \textsc{nyt-shuf}, we calculate the correlation between ranking average performance across experimental settings by the F1 score and by value of the difficulty metric (or expected ranking for group count/size). This was then averaged over all of the subsets to get their average correlation. The results of GPT-4 and GPT-3.5-Turbo are shown in Table~\ref{tab:background-averaged-bucket-and-correlation-results}.

\begin{table}[]
\centering
\tiny
\begin{tabular}{@{}c|cccc|c@{}}
\toprule
\textbf{Models}                                                                     & \multicolumn{4}{c|}{\textbf{Average F1 Score}}                            & \textbf{\begin{tabular}[c]{@{}c@{}}Average\\ Correlation\end{tabular}} \\ \midrule
\textit{Group Size}                                                                 & 2               & 3              & 4              & \multicolumn{1}{l|}{} &                                                                        \\ \midrule
GPT-3.5-Turbo                                                                       & 0.881           & \textbf{0.890} & 0.884          &                       & 0.250                                                                  \\
GPT-4                                                                               & 0.925           & 0.940          & \textbf{0.949} &                       & 0.750                                                                  \\ \midrule
\textit{Group Count}                                                                & 2               & 3              & 4              &                       &                                                                        \\ \midrule
GPT-3.5-Turbo                                                                       & \textbf{0.930}  & 0.883          & 0.842          &                       & 1.000                                                                  \\
GPT-4                                                                               & \textbf{0.962}  & 0.940          & 0.912          &                       & 1.000                                                                  \\ \midrule
\textit{\begin{tabular}[c]{@{}c@{}}Adjusted Rand\\ Index Range\end{tabular}}        & {[}-0.5, 0.0{]} & (0.0, 0.5{]}   & (0.5, 1.0{]}   &                       &                                                                        \\ \midrule
GPT-3.5-Turbo                                                                       & 0.924           & 0.927          & \textbf{0.951} & \textbf{}             & 0.900                                                                  \\
GPT-4                                                                               & 0.969           & 0.974          & \textbf{0.990} & \textbf{}             & 0.900                                                                  \\ \midrule
\textit{\begin{tabular}[c]{@{}c@{}}Candidate Proposal\\ Overlap Range\end{tabular}} & {[}0.0, 0.75{]} & (0.75, 1.5{]}  & (1.5, 2.25{]}  & (2.25, 3.0{]}         &                                                                        \\ \midrule
GPT-3.5-Turbo                                                                       & \textbf{0.932}  & 0.873          & 0.813          & 0.831                 & -0.800                                                                 \\
GPT-4                                                                               & \textbf{0.977}  & 0.923          & 0.863          & 0.859                 & -0.800                                                                 \\ \bottomrule
\end{tabular}
\caption{Average F1 results in games within buckets of specific difficulty metrics. Additionally, the ranking correlation of ranking average experimental setting performance by F1, and by the difficulty metric ranking, is provided.}
\label{tab:background-averaged-bucket-and-correlation-results}
\end{table}

\paragraph{Group Size \& Group Count} 
Contrary to our hypothesis for group size, as shown in Table \ref{tab:background-averaged-bucket-and-correlation-results}, the observed patterns are inconsistent. This is reflected by the large average correlation score of $0.750$ for GPT-4, contrasted with a very low score of $0.250$ for GPT-3.5-Turbo, indicating that game performance does not consistently correlate with variations in word grouping size.
In contrast, varying the number of groups reveals clear patterns in model performance. This is reflected in the correlation value, with both models boasting perfect correlations of $1.000$, validating our hypothesis for group count.

\paragraph{Adjusted Rand Index}
We compare ranking experimental settings by model performance to ranking by the ARI scores. As shown in Table \ref{tab:background-averaged-bucket-and-correlation-results}, both GPT-4 and GPT-3.5-Turbo exhibit near-perfect correlation coefficients of $0.900$. We conclude that increased inter-word similarity within a group increases performance, confirming our initial hypothesis.

\paragraph{Word Overlap Assessment}
We follow the same experimental setup to calculate the average correlation, as done in the clustering experiment. As shown in Table \ref{tab:background-averaged-bucket-and-correlation-results}, word overlap proves to be a significant predictor of game difficulty, with both models exhibiting a near-perfect correlation of $-0.800$, with performance trending as expected, decreasing as word overlap increases. Consequently, we conclude that word overlap is another indicator of game difficulty.

\subsection{Difficulty-aware Results}
\label{subsec:difficulty-results}
\begin{table*}[]
\tiny
\centering
\begin{tabular}{@{}c|ccccccllll@{}}
\toprule
\textbf{Models}       & \multicolumn{10}{c}{\textbf{Average F1 Score}}                                                                                                                                    \\ \midrule
\textit{Backgrounds}  & \multicolumn{1}{l}{en} & es    & es-en & zh             & zh-en          & hi             & \multicolumn{1}{c}{hi-en} & ar                      & ar-en          & nyt-shuf       \\ \midrule
\textit{Metric/Range} & \multicolumn{10}{c}{C1: ARI {[}-1.0, 0.0{]}, Word Overlap {[}0.0, 1.5{]}}                                                                                                         \\ \midrule
\textit{Game Count}   & 32                     & 31    & 10    & 1              & 13             & 164            & 7                         & 27                      & 38             & 37             \\ \midrule
GPT-3.5-Turbo         & 0.883                  & 0.790 & 0.838 & 0.875          & 0.755          & 0.901          & \textbf{0.982}            & 0.791                   & 0.911          & 0.581          \\
GPT-4                 & 0.947                  & 0.868 & 0.975 & 0.875          & 0.889          & \textbf{0.976} & 0.946                     & 0.914                   & 0.964          & 0.794          \\ \midrule
\textit{Metric/Range} & \multicolumn{10}{c}{C2: ARI {[}-1.0, 0.0{]}, Word Overlap (1.5, 3.0{]}}                                                                                                           \\ \midrule
\textit{Game Count}   & 0                      & 0     & 5     & 0              & 2              & 4              & 3                         & 0                       & 4              & 5              \\ \midrule
GPT-3.5-Turbo         & N/A                    & N/A   & 0.788 & N/A            & \textbf{0.938} & 0.766          & 0.875                     & N/A                     & 0.828          & 0.612          \\
GPT-4                 & N/A                    & N/A   & 0.850 & N/A            & \textbf{1.000} & 0.906          & 0.917                     & \multicolumn{1}{c}{N/A} & 0.844          & 0.750          \\ \midrule
\textit{Metric/Range} & \multicolumn{10}{c}{C3: ARI (0.0, 1.0{]}, Word Overlap {[}0.0, 1.5{]}}                                                                                                            \\ \midrule
\textit{Game Count}   & 217                    & 231   & 215   & 295            & 231            & 120            & 259                       & 270                     & 226            & 181            \\ \midrule
GPT-3.5-Turbo         & 0.875                  & 0.835 & 0.864 & 0.927          & 0.851          & 0.872          & \textbf{0.956}            & 0.755                   & 0.921          & 0.618          \\
GPT-4                 & 0.948                  & 0.902 & 0.927 & 0.971          & 0.938          & 0.966          & \textbf{0.974}            & 0.873                   & 0.959          & 0.840          \\ \midrule
\textit{Metric/Range} & \multicolumn{10}{c}{C4: ARI (0.0, 1.0{]}, Word Overlap (1.5, 3.0{]}}                                                                                                              \\ \midrule
\textit{Game Count}   & 24                     & 21    & 27    & 4              & 26             & 7              & 10                        & 2                       & 15             & 44             \\ \midrule
GPT-3.5-Turbo         & 0.849                  & 0.824 & 0.829 & \textbf{0.938} & 0.846          & 0.848          & 0.831                     & 0.562                   & 0.871          & 0.578          \\
GPT-4                 & 0.898                  & 0.856 & 0.910 & \textbf{0.969} & 0.907          & 0.827          & 0.853                     & 0.875                   & 0.917          & 0.777          \\ \midrule
\textit{Metric/Range} & \multicolumn{10}{c}{C5: ARI (0.5, 1.0{]}, Word Overlap {[}0.0, 0.75{]}}                                                                                                           \\ \midrule
\textit{Game Count}   & 12                     & 4     & 14    & 50             & 14             & 0              & 56                        & 12                      & 10             & 4              \\ \cmidrule(l){2-11} 
GPT-3.5-Turbo         & \textbf{0.979}         & 0.906 & 0.911 & 0.956          & 0.848          & N/A            & 0.971                     & 0.807                   & 0.942          & 0.688          \\
GPT-4                 & 0.990                  & 0.969 & 0.964 & 0.978          & 0.920          & N/A            & 0.990                     & 0.922                   & \textbf{1.000} & \textbf{1.000}
\end{tabular}
\caption{The averaged F1 score for closed-source models across 4-word, 4-group game. Results are divided by 2 ranges for the ARI and word overlap scores, creating 4 combinations (C1-4 in the first 4 rows), and by each dataset background, excluding \textsc{nyt-seq}. The number of games for each dataset background for each combination is provided as Game Count. C5 is a more refined difficulty metric combination, representing the easiest cases.}
\label{tab:4g4w-canpropoverlap-bucket-background-results}
\end{table*}

After establishing effective difficulty measures, we report performance trends across different dataset backgrounds while controlling for game difficulty, as shown in Table~\ref{tab:4g4w-canpropoverlap-bucket-background-results}. The datasets are categorized into four difficulty combinations based on two metrics: ARI score (where higher values indicate easier games) and word overlap score (where higher values indicate harder games). All experiments were conducted on 4-word, 4-group games across 10 benchmark subsets. 

We can note that in general, more difficult groups lead to lower performance for both models. We see this clearly when comparing C3 and C4 (with an increase in word overlap), with a stark decline in performance across all models and all backgrounds, except for GPT-3.5-Turbo with zh and GPT-4 with ar. We additionally see the same general pattern when comparing C1 and C3 (with an increase in ARI), with an increase in performance across all models and nearly all backgrounds. It is important to note that the backgrounds where this pattern is strong in the opposite direction are \textsc{hi} and \textsc{ar}, and that the expected pattern is partially restored in the English-translated versions of these datasets. These backgrounds utilize non-Latin script text, where the pretrained embeddings may not be adequately capturing word meaning, and therefore, may impact the ARI calculation.

We also note that most games fall within the easiest difficulty groups, which is likely due to the random sampling approach. This method is less likely to generate \textsc{nyt-seq}-like games with higher word overlap between games, although it does not completely eliminate the possibility, as shown by some games still appearing in higher overlap ranges.

\subsection{Devising An Integrated Difficulty Metric}
\label{app:integrated_difficulty_metric}

Utilizing the correlation results from Appendix~\ref{app:additional_difficulty_metric_analysis}, we can create an integrated difficulty metric for a \ours{} game by considering group count, ARI, and word overlap together. For this, we:
\begin{itemize}
    \item Project each metric's value, considering the metric's range, onto a range of $[0,1]$. For group count we project from the range $[2, 4]$, for ARI we project from the range $[-1, 1]$, and for word overlap we project from the range $[0, 3]$.
    \item Utilizing the previously calculated correlation coefficients, we perform a weighted summation of the projected values, with each value weighted by its metric's respective correlation coefficient in the direction of difficulty. For group count this would be $1.000$ (as a higher group count is associated with more difficulty), for ARI this would be $-0.900$ (as a higher ARI is associated with less difficulty), and for word overlap this would be $0.800$ (as a higher word overlap is associated with more difficulty).
\end{itemize}
This results in an integrated difficulty metric within the range of $[-0.9, 1.8]$. This is finally projected into the range $[0, 1]$ as the final score.

Performing an identical correlation analysis with the integrated difficulty metric as done for ARI and word overlap yields a perfect correlation of $-1.000$, meaning that as the integrated difficulty metric value of a game increases, F1 performance on that game decreases.

\end{document}